\documentclass[sigconf]{acmart}

\usepackage[utf8]{inputenc} 
\usepackage[T1]{fontenc}    
\usepackage{hyperref}       
\usepackage{url}            
\usepackage{booktabs}       
\usepackage{amsfonts}       
\usepackage{microtype}      
\usepackage{xcolor}
\usepackage{graphicx}
\usepackage{amsmath}
\usepackage{amsthm}
\usepackage{array}

\usepackage{amsfonts}
\usepackage{algorithmic}
\usepackage{textcomp}
\usepackage{booktabs}
\usepackage{enumitem}
\usepackage{verbatim}
\usepackage{subfigure}
\usepackage{soul}
\usepackage{multirow,multicol}
\usepackage{balance}
\usepackage{ifthen}
\usepackage{float}

\usepackage[linesnumbered,ruled,vlined]{algorithm2e}

\SetKwInput{KwInput}{Input}                
\SetKwInput{KwOutput}{Output}              

\AtBeginDocument{%
  \providecommand\BibTeX{{%
    \normalfont B\kern-0.5em{\scshape i\kern-0.25em b}\kern-0.8em\TeX}}}

\copyrightyear{2022}
\acmYear{2022}
\setcopyright{rightsretained}
\acmConference[CCS '22]{Proceedings of the 2022 ACM SIGSAC Conference on Computer and Communications Security}{November 7--11, 2022}{Los Angeles, CA, USA}
\acmBooktitle{Proceedings of the 2022 ACM SIGSAC Conference on Computer and Communications Security (CCS '22), November 7--11, 2022, Los Angeles, CA, USA}
\acmDOI{10.1145/3548606.3560573}
\acmISBN{978-1-4503-9450-5/22/11}

\usepackage{etoolbox}
\makeatletter
\patchcmd{\maketitle}{\@copyrightpermission}{
   \begin{minipage}{0.3\columnwidth}
     \href{https://creativecommons.org/licenses/by/4.0/}{\includegraphics[width=0.90\textwidth]{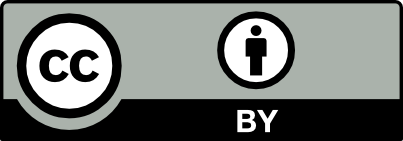}}
   \end{minipage}\hfill
   \begin{minipage}{0.7\columnwidth}
     \href{https://creativecommons.org/licenses/by/4.0/}{This work is licensed under a Creative Commons Attribution International 4.0 License.}
   \end{minipage}

   \vspace{5pt}
}{}{}

\makeatother

\begin{document}



\title{Feature Inference Attack on Shapley Values}\titlenote{This paper presents a revision of the previously published version, addressing an error discovered in the proof of Theorem 1. The main update involves replacing Theorem 1 with an empirical discussion. Note that Theorem 1 served primarily as an attack motivation, and its removal does not affect the proposed attack methods and their results.}


\author{Xinjian Luo}
\affiliation{%
 \institution{National University of Singapore}
 \country{}
}
\email{xinjluo@comp.nus.edu.sg}

\author{Yangfan Jiang}
\affiliation{%
 \institution{National University of Singapore}
 \country{}
}
\email{yangfan.jiang@comp.nus.edu.sg}

\author{Xiaokui Xiao}
\affiliation{%
 \institution{National University of Singapore}
 \country{}
}
\email{xkxiao@nus.edu.sg}

\settopmatter{printfolios=true}


\begin{abstract}

As a solution concept in cooperative game theory, Shapley value is highly recognized in model interpretability studies and widely adopted by the leading Machine Learning as a Service (MLaaS) providers, such as Google, Microsoft, and IBM.
However, as the Shapley value-based model interpretability methods have been thoroughly studied, few researchers consider the privacy risks incurred by Shapley values, despite that interpretability and privacy are two foundations of machine learning (ML) models.

In this paper, we investigate the privacy risks of Shapley value-based model interpretability methods using feature inference attacks: \textit{reconstructing the private model inputs based on their Shapley value explanations}. 
Specifically, we present two adversaries. The first adversary can reconstruct the private inputs by training an attack model based on an auxiliary dataset and black-box access to the model interpretability services. The second adversary, even without any background knowledge, can successfully reconstruct most of the private features by exploiting the local linear correlations between the model inputs and outputs.
We perform the proposed attacks on the leading MLaaS platforms, i.e., Google Cloud, Microsoft Azure, and IBM aix360.
The experimental results demonstrate the vulnerability of the state-of-the-art Shapley value-based model interpretability methods used in the leading MLaaS platforms and highlight the significance and  necessity of designing privacy-preserving model interpretability methods in future studies.
To our best knowledge, this is also the first work that investigates the privacy risks of Shapley values.
\end{abstract}

\clearpage
\pagenumbering{gobble}




\maketitle

\section{Introduction}\label{sec-intro}
Nowadays, machine learning (ML) models are being increasingly deployed into high-stakes domains for decision making, such as finance, criminal justice and employment~\cite{rudin2019stop, arya2019IBM, chen2018learning, datta2016algorithmic}. 
Although demonstrating impressive prediction performances, most deployed ML models behave like black boxes to the developers and practitioners because of their complicated structures, e.g., neural networks, ensemble models and kernel methods~\cite{chen2018learning, lundberg2017SHAP}. 
To foster trust in ML models, researchers design numerous ``explainable'' or ``interpretable'' methods to help humans understand the predictions of ML systems. In this paper, we focus on \textit{Shapley value-based local interpretability methods}, i.e., the class of methods that use Shapley value to explain individual model predictions.


Shapley value~\cite{shapley1953} is first proposed to distribute the surplus among a coalition of players in cooperative game theory. Since it theoretically satisfies a collection of desirable properties, i.e., symmetry, linearity, efficiency and null player~\cite{shapley1953}, Shapley value is widely recognized and adopted as the model interpretability method in both academia~\cite{covert2020SAGEGlobalShapely, lundberg2017SHAP, datta2016algorithmic, chen2018shapley, maleki2013bounding, vstrumbelj2014explaining, chen2018learning} and the leading MLaaS platforms, such as  Google Cloud, Microsoft Azure and IBM aix360~\cite{Google, Microsoft, IBM}.
But with the increasing deployment of ML models in the high-stakes domains, few studies related to model interpretability consider the privacy issues, despite that privacy and interpretability are two fundamental properties demanded by regulations~\cite{goodman2017european, selbst2018meaningful}.

\begin{figure*}[t]
\centering
\includegraphics[width=0.8\textwidth]{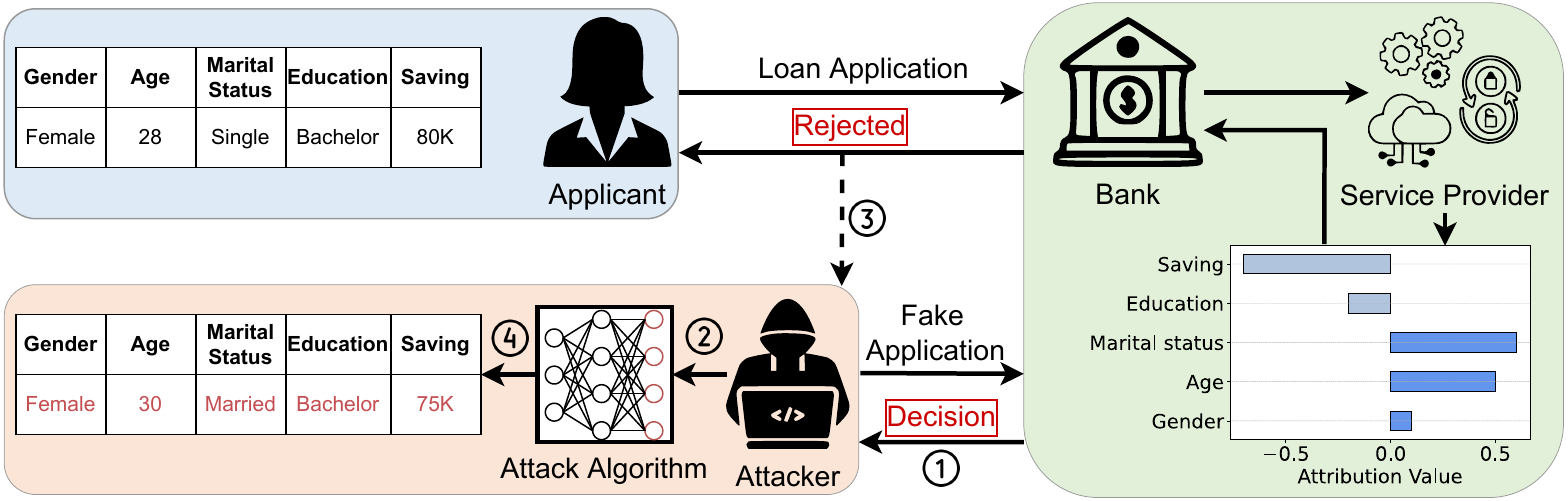}
\vspace{-2mm}
\caption{Attack framework based on explanation reports: (1) the attacker sends fake queries to ML platforms and receives decisions with explanations; (2) based on the model inputs and explanations, the attacker designs a feature inference algorithm; (3) the explanation reports from the target customers may be obtained by the attacker; (4) from these explanations, the attacker can reconstruct the corresponding private features via the attack algorithm. }
\label{fig-attack-illustration}
\vspace{-3mm}
\end{figure*}

\vspace{0.5mm}
\noindent
\textbf{Motivation}. Interpretability is a significant aspect of ML models. 
Specifically, ``the right to explanation'' has been included in the European Union General Data Protection Regulation (GDPR)~\cite{kim2018informational, goodman2017european, selbst2018meaningful}.
Model interpretability is especially critical in the high-stakes domains. For example, in a data-driven loan platform supported by a financial institution, a customer who submits a loan application will receive an approval decision recommended by an ML model~\cite{arya2019IBM, ribeiro2016LIME}. If the loan application is denied, the customer has the right to demand an explanation report behind the denial from the ML platform, as shown in Fig.~\ref{fig-attack-illustration}.
It is worth noting that unlike the private model inputs which are classified into highly sensitive information in real-world applications, the explanation reports\footnote{\label{footnote-sep-input-exp}
{
Note that whether or not to include the private features in the explanation reports depends on the practitioners and target users. For example, QII~\cite{datta2016algorithmic} and the demo on Google AI Platform~\cite{Google} only illustrate Shapley values in explanation reports. Meanwhile, the demos on IBM x360~\cite{IBM} and SHAP library~\cite{lundberg2017SHAP} include both the explanations and private features. In this paper, we focus on the first case.
}} can be available to other personas except the customers and loan officers. 
{We give some examples of explanation releases as follows. For companies, the explanations can help evaluate model bias and correctness, e.g., identifying dataset shift where training data is different from test data~\cite{ribeiro2016LIME}, and understanding where the model is more or less confident for further improving its performance~\cite{arya2019IBM}. In these cases, the company needs to transmit model explanations to ML service providers (which could be the third party) for analysis~\cite{arya2019IBM}. For customers, the explanations could be shared on social media and obtained by the adversary~\cite{zhao2021exploiting}. Because the privacy risks in Shapley values are not well studied, both the companies and customers may underestimate the sensitivity of model explanations and transmit them in plaintext.}
%
%
Therefore, we are inspired to ask: \textit{are there any privacy risks in the model explanations generated via Shapley value?}
Although some studies exploit membership inference~\cite{shokri2021privacy} and model extraction~\cite{aivodji2020model, milli2019model} on model explanations, all of them focus on the gradient-based heuristic methods instead of Shapley values, and the information inferred by these attacks has little relation to the ground-truth model inputs.
In this paper, we demonstrate the vulnerability of Shapley values by feature inference attack, i.e., reconstructing the private features of model inputs based on their Shapley value explanations.

\vspace{0.5mm}
\noindent
\textbf{Challenges}.
{There are three main challenges when performing feature inference on Shapley values.
\textit{First} is how to develop an in-depth analysis on exactly how much privacy could be leaked by Shapley values. Since the privacy analyses in previous works~\cite{shokri2021privacy, aivodji2020model, milli2019model} mainly target on the gradient-based explanations which are totally different from Shapley values, we have to analyze the information flow from private inputs to their Shapley values from scratch.
\textit{Second} is how to design generalized attacks applicable to both the original Shapley values and its variants~\cite{maleki2013bounding, lundberg2017SHAP}. For computational efficiency, the sampling methods, e.g., Shapley sampling values~\cite{maleki2013bounding} and SHAP~\cite{lundberg2017SHAP}, are preferred in real-world applications.
Considering that the explanations produced by sampling methods could  randomly and greatly deviate from the ground-truth Shapley values~\cite{lundberg2017SHAP}, the accuracies of feature inference attacks could be significantly impacted by these unstable explanations, because feature inference focuses on recovering private information with the finest granularity, i.e., the feature values of the model inputs.
%
%
\textit{Third} is how to accurately infer private features via a limited number of queries to the MLaaS platforms. Because different from previous adversaries~\cite{shokri2021privacy, melis2019exploiting} who have unlimited access to the target models, the adversary in the real-world setting needs to access the ML service in a pay-per-query pattern~\cite{wu2020privacy}.
A brute force method that sends unlimited queries to the explanation platform may help accurately estimate the private features, but can be easily detected by the service provider and bring prohibitive financial and time costs to the adversaries.
%
%
An attack algorithm that accurately reconstructs private features with a bounded error from a small number of queries is necessary in real-world applications.}

\vspace{0.5mm}
\noindent
\textbf{Contributions}.
In this paper, to design attacks that can be used on Shapley sampling values with unknown sampling errors, we first analyze the connections between model inputs and the associated explanations in an information-theoretical perspective, and show that the information of private features is contained in its Shapley values as long as the variance of sampling errors is smaller than that of the Shapley values.  
Then, we reveal the vulnerability of Shapley value-based model interpretability methods by two feature inference attacks. These attacks can be applied to both the original Shapley values and Shapley sampling values, and a limited number of queries (e.g., 100) to the ML services can enable the adversary to accurately reconstruct the important private features corresponding to the target explanation reports. 
Specifically, we present two adversaries. The \textit{first} adversary follows a similar setting to current studies~\cite{shokri2021privacy, salem2018ml, ganju2018property, melis2019exploiting} in which the adversary owns an auxiliary dataset of model inputs. 
Based on the auxiliary dataset and black-box access to the explanation service, the adversary can train an attack model by empirically minimizing the sampling errors.
The \textit{second} adversary is presented with a relaxed assumption of the first one, who has only black-box access to the ML platform and no background knowledge about the private features.
Feature inference is harder to achieve for this adversary because the search space of private features is infinite and the errors of the reconstructed features, if any, are hard to be estimated without any auxiliary knowledge.
Nevertheless, we analyze the correlations between model inputs and outputs as well as the Shapley values, and find that the local linearity of model decision boundaries for some important features can be passed to Shapley values. Accordingly, we propose an attack algorithm by first approximating the black-box model with a generalized additive model and then estimating the private features via linear interpolations based on a set of randomly generated model inputs. The estimation error of the second adversary can be further bounded by Chebyshev's inequality and Hoeffding's inequality.
In the experiments performed on three leading model interpretability platforms (Google Cloud~\cite{Google}, Microsoft Azure~\cite{Microsoft} and IBM aix360~\cite{IBM}) with six real-world datasets and four black-box models, we show that with only 100 queries, the first adversary can reconstruct the private features with an averaging $10\%$ deviation from the ground truth features, and the second adversary can reconstruct at least $30\%$ features with an averaging $14\%$ deviation. Our contributions are summarized as follows:
\begin{itemize}[topsep=2mm]
  \item We formally investigate the feasibility of feature inference attacks on Shapley value-based model interpretability methods. We show that except ``the right to explanation'', ``the right to be forgotten of explanations'' should also be seriously considered and regulated. To our best knowledge, this is also the first study that focuses on the privacy risks of Shapley values. 
  \item We propose two adversaries with different assumptions. The first adversary can reconstruct the private features via an attack model trained on an auxiliary dataset, while the second adversary can infer the important features  by analyzing the local linearity of model decision boundaries, without the assistance of any background knowledge.
  \item We perform the proposed attacks on three leading model interpretability platforms with six real-world datasets, three synthesis datasets and four black-box models. The experiments demonstrate the effectiveness of our attacks. We analyze several defense mechanisms against the proposed attacks and highlight the necessity for developing privacy-preserving model interpretability methods.
\end{itemize}

\section{Preliminary}

\subsection{Machine Learning}\label{subsec-preliminary-ml}
We focus on the supervised machine learning tasks. Let $\mathcal{D}=\{(\boldsymbol{x}^t, \boldsymbol{y}^t)\in \mathcal{X}\times \mathcal{Y}\}^m_{t=1}$ denote a training dataset with $m$ samples, where $\mathcal{X}\subseteq \mathbb{R}^n$ denotes the input space with $n$ features, and $\mathcal{Y} \subseteq \mathbb{R}^c$ denotes the output space with $c$ classes. Machine learning aims to train a model $f$ with parameters $\boldsymbol{\theta}$ such that the following loss function is minimized:
\begin{equation}\label{eq-ml}
  \min_{\boldsymbol{\theta}}\frac{1}{m}\sum_{t=1}^{m}\ell(f(\boldsymbol{x}^t;\boldsymbol{\theta}), \boldsymbol{y}^t)+\Omega(\boldsymbol{\theta}),
\end{equation}
where $\Omega(\boldsymbol{\theta})$ denotes the regularization term for preventing the model from overfitting.

\vspace{0.5mm}\noindent
\textbf{Neural Networks}. Neural network (NN) models consist of one input layer, one output layer and multiple hidden layers, where non-linear transformations, such as ReLU and Sigmoid, are used as the activation functions after the hidden layers. Although demonstrating impressive performance over the last few years, NN is less trustable compared to the traditional linear models because the information flow in NN is hard to be inspected~\cite{ancona2017towards}.

\vspace{0.5mm}\noindent
\textbf{Ensemble Models}. Ensemble methods can improve the prediction performance by first training multiple models and then combining their outputs as the final prediction~\cite{sagi2018ensemble}. Two of the most popular ensemble models are Random Forest (RF)~\cite{breiman2001random} and Gradient Boosted Decision Trees (GBDT)~\cite{friedman2001greedy}. Although both of them are composed of multiple decision trees, the training phases of these two models are mostly different. Each tree in RF is independently trained based on a subset of randomly selected samples and features, whereas the trees in GBDT are connected and one successive tree needs to predict the residuals of the preceding trees. Because the predictions of the ensemble models are aggregated from the outputs of multiple trees (e.g., voting in RF and addition in GBDT), it is still a difficult task to interpret these ensemble ouputs~\cite{sagi2018ensemble}.

\vspace{0.5mm}\noindent
\textbf{Kernel Methods}. Kernel methods are mainly used to recognize the nonlinear patterns in the training datasets~\cite{pilario2019review}. The main idea is to first project the input features into a high-dimensional space via non-linear transformations and then train linear models in that space. One of the most famous kernel models is the support-vector machine (SVM) with kernel trick~\cite{boser1992training}. Note that kernel SVMs learn data patterns of the transformed features in the high-dimensional space instead of the original features, which makes its predictions hard to be interpreted~\cite{ratsch2006learning}.

\subsection{Shapley Value}\label{subsec-shapley-value}

Shapley value~\cite{shapley1953} is a widely recognized method for interpreting model outputs in machine learning~\cite{covert2020SAGEGlobalShapely, lundberg2017SHAP, datta2016algorithmic}.
In this paper, we focus on \emph{local interpretability}: for an input sample $\boldsymbol{x}=\{x_i\}^n_{i=1}$ with $n$ features and a specific class in the model outputs, Shapley value computes a vector $\boldsymbol{s}=\{s_{i}\}^n_{i=1}$ in which each element $s_i$ denotes the influence of the corresponding feature $x_i$ on the target class of model outputs. Now we introduce how to compute the importance value ${s}_i$ for feature $x_i$.

Let $N=\{1, \cdots, n\}$ be the index set of all features, $S\subseteq N$ be a collection of features, and $\boldsymbol{x}^0=\{x^0_i\}^n_{i=1}$ be a reference sample. A sample $\boldsymbol{x}_{[S]}$ is composed as follows: $\forall j\in N$, $(\boldsymbol{x}_{[S]})_j=x_j$ if $j\in S$ and $(\boldsymbol{x}_{[S]})_j=x^0_j$ if $j\in N\backslash S$.
Now given the feature indexes $N$ and a trained model $f$, the Shapley value of feature $i$ is defined by:
\begin{equation}\label{eq-shap}
  {s}_i=\frac{1}{n}\sum_{S \subseteq N \backslash \{i\}}
    \frac{ f(\boldsymbol{x}_{[S\cup \{i\}]})- f(\boldsymbol{x}_{[S]})}{{n-1 \choose |S|}}.
\end{equation}
Note that $m_{i}(S, \boldsymbol{x})=f(\boldsymbol{x}_{[S\cup \{i\}]})- f(\boldsymbol{x}_{[S]})$ represents the marginal contribution of feature $i$ to the feature set $S$. For example, suppose $\boldsymbol{x}^0 = \begin{bmatrix} 3& 9& 2& 8\end{bmatrix}$, $\boldsymbol{x}=\begin{bmatrix} \textbf{6} & \textbf{0} &  \textbf{3} & \textbf{4} \end{bmatrix}$ and $N=\{1, 2, 3, 4\}$, we compute the marginal contribution of feature $i=1$ to the feature set $S=\{2, 3\}$ by
$f(\boldsymbol{x}_{[\{1, 2, 3\}]}) - f(\boldsymbol{x}_{[\{2, 3\}]})=f\left( \begin{bmatrix} \boldsymbol{6}& \boldsymbol{0}& \boldsymbol{3}& 8  \end{bmatrix}\right)-f\left(\begin{bmatrix}  3& \boldsymbol{0}& \boldsymbol{3}& 8  \end{bmatrix} \right)$. 
Mathematically, we can interpret ${s}_i$ as the expectation of marginal contributions of feature $i$ to all possible sets composed by the other features $S\subseteq N\backslash \{i\}$~\cite{fujimoto2006axiomatic,chen2018shapley}.

\vspace{0.5mm}\noindent
\textbf{Shapley sampling value}.
The $O(2^{n})$ complexity of Eq.~\ref{eq-shap} is computationally prohibitive for most ML applications, thus most studies~\cite{datta2016algorithmic,lundberg2017SHAP} and MLaaS platforms~\cite{Google} use a sampling method~\cite{maleki2013bounding} to compute the approximate Shapley values: let $r_{m_{i}}$ be the range of the marginal contributions of feature $i$, and $\{P^N_k\}^{\upsilon}_{k=1}$ be $\upsilon$ permutations of $N$. For each permutation $P^N_k$, $S$ denotes the set of features that precede $i$.
By setting $\upsilon\geq \left\lceil\frac{\ln(\frac{2}{\delta})r_{m_{i}}^2}{2\epsilon^2} \right\rceil$, we can compute the approximate Shapley values $\hat{{s}}_i$ by Eq.~\ref{eq-shap} such that
\begin{equation}\label{eq-shap-sampling}
  \text{Pr}(|\hat{{s}}_i - {s}_i |\geq \epsilon) \leq \delta.
\end{equation}
Here $\epsilon>0$ denotes the estimation error.

\section{Problem Statement}\label{sec-problem}
\vspace{0.5mm}
\textbf{System Model.}
In this paper, we consider a system model in which a commercial institution trains a black-box model $f$ based on a sensitive dataset $\mathcal{X}_{train}$ and deploy it on an ML-as-a-service (MLaaS)~\cite{tramer2016stealing} platform, such as Google Cloud~\cite{Google} and Microsoft Azure~\cite{Microsoft}. Users can access the model based on a pay-per-query pattern~\cite{wu2020privacy}.
Specifically, a user can send a private sample $\boldsymbol{x}=\{x_i\}^n_{i=1}$ to the service provider and obtain a prediction vector $\hat{\boldsymbol{y}} =\{\hat{y}_i\}^c_{i=1}$ with an explanation vector $\boldsymbol{s}=\{s_{i}\}^n_{i=1}$ \textit{w.r.t.} a specific class, as shown in Fig.~\ref{fig-attack-illustration}.
Note that the service provider can also return $c$ explanation vectors, each of which corresponds to a class, but returning one vector $\boldsymbol{s}$ \textit{w.r.t.} a predefined class is the default setting in related studies~\cite{shokri2021privacy} and MLaaS platforms, e.g., Google Cloud~\cite{Google}. Thus, for practicability and without loss of generality, we consider one explanation vector \textit{w.r.t.} a specific class in this paper.
We summarize the frequently used notations in Tab.~\ref{table-notations} for reference.

\vspace{0.5mm}\noindent
\textbf{Attack Model.}
The previous studies in ML privacy~\cite{shokri2021privacy, melis2019exploiting, salem2018ml, nasr2019comprehensive} assume that the adversary can collect an auxiliary dataset $\mathcal{X}_{aux}$ which follows the same underlying distribution of the target sample $\boldsymbol{x}$. For example, a bank can synthesize a real-world dataset after obtaining the feature names on the explanation report from its competitors. In this paper, we study two adversaries by relaxing the assumptions of previous studies. The \textit{first} adversary follows the setting in~\cite{shokri2021privacy, melis2019exploiting, salem2018ml, nasr2019comprehensive}, i.e., the attacker aims to infer the private sample $\boldsymbol{x}$ based on the associated explanation vector $\boldsymbol{s}$, an auxiliary dataset $\mathcal{X}_{aux}$ and black-box access to the prediction model $f$ on the MLaaS platforms: $ \hat{\boldsymbol{x}} = \mathcal{A}_1(\boldsymbol{s}, \mathcal{X}_{aux}, f)$,
where $\hat{\boldsymbol{x}}$ denotes the reconstructed values of $\boldsymbol{x}$.
For the \textit{second} adversary, we use a more restricted but practical setting that the attacker has only black-box access to the ML services, and no background knowledge about the target sample $\boldsymbol{x}$ is available: $\hat{\boldsymbol{x}} = \mathcal{A}_2(\boldsymbol{s}, f)$.

\begin{table}[t]
\centering
\caption{Summary of notations}
\vspace{-2mm}
\begin{tabular}{  l  l }
\toprule
Notation & Description \\
\toprule
$f$ & the black-box model deployed on an MLaaS platform \\
$\boldsymbol{x}$ & the target sample to be reconstructed\\
$n$ & the number of features in $\boldsymbol{x}$\\
$\hat{\boldsymbol{y}}$ & the output of the black-box model, i.e., $f(\boldsymbol{x})$ \\ 
$c$ & the number of classes in $\hat{\boldsymbol{y}}$ \\
$\boldsymbol{s}$ & the Shapley values of $\boldsymbol{x}$ \\
$\boldsymbol{x}^0$ & the reference sample for computing $\boldsymbol{s}$\\
$x^j_i$ & the $i$-th feature of the $j$-th sample $\boldsymbol{x}^j$  \\
\bottomrule
\end{tabular}
\label{table-notations}
\vspace{-5mm}
\end{table}
\section{Feature Inference Attacks Based on Shapley Values}
In this section, we present the proposed feature inference attacks under different settings. Based on an auxiliary dataset, the first adversary can first learn a regression model with empirical risk minimization, then map the explanation vector $\boldsymbol{s}$ into the private feature space and accordingly obtain the estimation $\hat{\boldsymbol{x}}$ of $\boldsymbol{x}$. With only black-box access to the MLaaS platform, the second adversary can accurately reconstruct the important features of $\boldsymbol{x}$ by exploiting the linearity of local model decision boundaries.

\subsection{Adversary 1: Feature Inference Attack with an Auxiliary Dataset}\label{subsec-attack-1}
\subsubsection{Motivation}\label{subsec-a1-motivation}
Given an input sample $\boldsymbol{x}=\{x_i\}^n_{i=1}$, a reference sample $\boldsymbol{x}^0$ and the access to a black-box model $f$, we can simplify the computation of Shapley value in  Eq.~\ref{eq-shap} as follows:
\begin{equation}\label{eq-simply-shap}
  s_i=h(x_i;x_1,\cdots,x_{i-1},x_{i+1},\cdots,x_n,\boldsymbol{x}^0) + \epsilon,
\end{equation}
where $h=g\circ f$ is the composition of the black-box model $f$ and a linear transformation (i.e., the mathematical expectation) $g$ on $f$. $\epsilon$ denotes the random noise caused by the sampling phase (Eq.~\ref{eq-shap-sampling}).

Now we characterize the dependence between $s_i$ and $x_i$ via a popular information metric called \textit{Mutual Information}~\cite{chen2018shapley, chen2018learning}:
\begin{equation}\label{eq-mutual-info}
  I(x_i; s_i) = \int ds_i \int dx_i P(x_i, s_i)\left[\log \frac{P(x_i, s_i)}{P(x_i)P(s_i)} \right].
\end{equation}
To reach a perfect security level of Eq.~\ref{eq-simply-shap}, we need to make $I(x_i; s_i)=0$ such that the adversary can not obtain any useful information of $x_i$ by observing $s_i$~\cite{yeung2008information}.

For simplicity, we denote the variance and mean of a variable $x$ by $\sigma^2_x$ and $\mu_x$, respectively. We assume the noise $\epsilon$ caused by random sampling and the Shapley value $s_i$ computed by Eq.~\ref{eq-simply-shap} follow Gaussian distributions with mean 0 and some variances\footnote{\label{footnote-gaussian}
{We have conducted a $\chi^2$
  goodness-of-fit test on the Shapley values generated on the Google platform from two datasets: Bank~\cite{bank} and Diabetes~\cite{diabetes}, where $H_0$
  is ``Gaussian density (theoretical) distribution can well fit the density distribution of the observed Shapley values''. Test results show that for Shapley values generated from Bank and Diabetes, the divergence statistics are 3.84 and 2.48, and $p$ values are 0.998 and 0.999. Therefore, under $p\gg 0.05$, we can not reject $H_0$.}} 
because of the central limit theorem~\cite{maleki2013bounding, ancona2019explaining, covert2021improving}, thus
\begin{equation}\label{eq-gaussian-ps}
  P(s_i)=\frac{1}{\sqrt{2\pi \sigma^2_{s_i}} } \exp \left[ -\frac{s^2_i}{2\sigma^2_{s_i}} \right].
\end{equation}

Note that because $\epsilon$ is Gaussian noise and $\epsilon = s_i - h(x_i)$, then given $x_i$, $s_i$ also follows a Gaussian distribution $s_i|x_i\sim N(\mu_{s_i|x_i}, \sigma^2_{s_i|x_i})$, where $\mu_{s_i|x_i}=h(x_i)$ and $\sigma_{s_i|x_i}=\sigma_{\epsilon}$. Specifically,
\begin{equation}\label{eq-gaussian-s-on-x}
  P(s_i|x_i)=\frac{1}{\sqrt{2\pi \sigma^2_{\epsilon}} } \exp \left[ -\frac{(s_i-h(x_i)   )^2}{2\sigma^2_{\epsilon}} \right].
\end{equation}

 Then we can rewrite Eq.~\ref{eq-mutual-info} as follows:

  \begin{align}
    I(x_i; s_i) = & \int ds_i \int dx_i P(x_i, s_i)\left[\log \frac{P(s_i | x_i)}{P(s_i)} \right] \label{eq-mutual-information-duction-1} \\
    = & \frac{1}{\ln 2} \mathbb{E}\left[\ln \frac{\sqrt{2\pi \sigma^2_{s_i}}}{\sqrt{2\pi \sigma^2_{\epsilon}} } - \frac{(s_i-h(x_i))^2}{2\sigma^2_{\epsilon}} + \frac{s^2_i}{2\sigma^2_{s_i}}  \right] \label{eq-mutual-information-duction-2} \\
    {=} & \frac{1}{\ln 2} \mathbb{E}\left[\ln \sqrt{\frac{\sigma^2_{s_i}}{\sigma^2_{\epsilon}} } - \frac{1}{2} + \frac{1}{2} \right] \label{eq-mutual-information-duction-3} \\
    = & \frac{1}{2}\mathbb{E}\left[\log \frac{\sigma^2_{s_i}}{\sigma^2_{\epsilon}}  \right] \label{eq-mutual-information-duction-4},
  \end{align}
where Eq.~\ref{eq-mutual-information-duction-3} follows because for the second term of Eq.~\ref{eq-mutual-information-duction-2}, $\epsilon=s_i-h(x_i)$ and $\mathbb{E}(\epsilon^2)=\sigma^2_{\epsilon}+\mu_{\epsilon}=\sigma^2_{\epsilon}$, making the numerator cancel with the denominator, and so does for the third term.

Now to achieve a perfect level of security, we need to make $I(x_i;s_i)=0$, i.e., $\sigma^2_{s_i}=\sigma^2_{\epsilon}$ in Eq.~\ref{eq-mutual-information-duction-4}. But to obtain a reasonable interpretability $s_i$ of $x_i$, we need to minimize the noise component~\cite{chen2018learning}, which means that $\sigma^2_{s_i}>\sigma^2_{\epsilon}$ is necessary from the perspective of practicability. In summary, an adversary can always infer useful information of $x_i$ from the explanation report as long as the Shapley values are meaningful.

\subsubsection{The Attack with Auxiliary Datasets}
Previous studies, such as membership inference~\cite{shokri2021privacy, salem2018ml} and property inference~\cite{ganju2018property, melis2019exploiting}, assume that the adversary owns an auxiliary dataset $\mathcal{X}_{aux}$ following the same underlying distribution of the target samples. This assumption is reasonable because the competing companies own datasets that have the same features but different customer bases~\cite{kairouz2021advances}. The users of the MLaaS platforms can also collect a dataset $\mathcal{X}_{aux}$ by collusion~\cite{mohassel2017secureml}. In this section, we follow this assumption and show that the adversary can reconstruct the target samples by training a regression model on $\mathcal{X}_{aux}$.

Based on the analysis in Section~\ref{subsec-a1-motivation}, we know that the explanation vector $\boldsymbol{s}$ is related to the private sample $\boldsymbol{x}$. Although the intensity of the noise $\epsilon$ is unknown, the adversary can still learn a model to reconstruct $\boldsymbol{x}$ by minimizing the noise component. Specifically, the adversary can send prediction queries for all $\boldsymbol{x}_{aux}\in \mathcal{X}_{aux}$ to the MLaaS platform and obtain the corresponding explanation set $\boldsymbol{S}_{aux}$.
Now we rewrite Eq.~\ref{eq-simply-shap} as
\begin{equation}\label{eq-simply-shap-vector}
  \boldsymbol{s}=\phi(\boldsymbol{x};\boldsymbol{x}^0).
\end{equation}
The task becomes learning a hypothesis $\psi: \mathcal{S}_{aux} \rightarrow \mathcal{X}_{aux}$ that serves as the attack model. Here, $\mathcal{X}_{aux}\subseteq \mathbb{R}^n$ represents a real-world dataset with finite cardinality. Additionally, both $\boldsymbol{x}$ and $\boldsymbol{s}$ are real-valued vectors, and we have $|\mathcal{X}_{aux}|=|\mathcal{S}_{aux}|$. 
The key observation is that our experiments exhibit a one-to-one correspondence between $\mathcal{X}_{aux}$ and $\mathcal{S}_{aux}$. This correspondence enables us to learn a hypothesis $\psi: \mathcal{S}_{aux} \rightarrow \mathcal{X}_{aux}$ to model the relations between input samples and Shapley values. We can then use $\psi$ to produce the target $\boldsymbol{x}$ from an unknown $\boldsymbol{s}$. 
Note that collisions may occur during the generation of $\mathcal{S}_{aux}$ from $\mathcal{X}_{aux}$; that is, multiple input samples might result in the same Shapley vector. However, such collisions are infrequent and have not been observed in our experiments to date. When collisions do occur, we can address them by removing the affected $(\boldsymbol{x}, \boldsymbol{s})$ pairs before learning the hypothesis $\psi$.

\begin{figure}[t]
\centering
\includegraphics[width=.6\columnwidth]{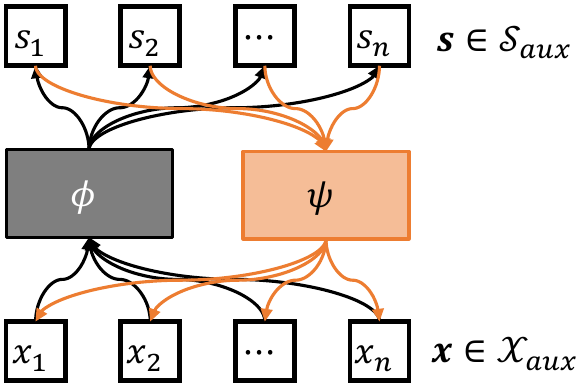}
\caption{Overview of the attack with an auxiliary dataset. $\psi$ is the attack model.}
\label{fig-attack1-overview}
\vspace{-3mm}
\end{figure}

It is also important to note that learning a hypothesis $\psi: \mathcal{S}_{aux} \rightarrow \mathcal{X}_{aux}$ does not imply that the function $\phi: \mathcal{X}_{aux} \rightarrow \mathcal{S}_{aux}$ is invertible. The invertibility of  $\phi$ depends on the properties of the underlying models, which is not guaranteed for most model types.
Nevertheless, the size of the auxiliary dataset $|\mathcal{X}_{aux}|$ is always finite in real-world applications, and the adversary can typically learn a hypothesis $\psi: \mathcal{S}_{aux} \rightarrow \mathcal{X}_{aux}$ by minimizing the empirical risk as shown in Fig.~\ref{fig-attack1-overview}:
\begin{equation}\label{eq-mini-risk}
  \min_{\psi\in\mathcal{H}} \frac{1}{|\mathcal{X}_{aux}|} \sum L(\psi(\mathcal{S}_{aux}), \mathcal{X}_{aux}) + \Omega(\psi),
\end{equation}
where $L$ denotes the loss function, $\mathcal{H}$ denotes a set of hypotheses and $\Omega$ is the regularization term.
After learning a model $\psi$, the adversary can use it to reconstruct a private sample $\boldsymbol{x}$ based on its Shapley values $\boldsymbol{s}$: $\hat{\boldsymbol{x}} = \psi (\boldsymbol{s})$.
Note that the regularization term in Eq.~\ref{eq-mini-risk} is significant to help the model focus on the information of $\boldsymbol{x}$ instead of the noise $\epsilon$ in $\boldsymbol{s}$. The attack with an auxiliary dataset is summarized in Algorithm~\ref{alg-attack-1}.

\begin{figure}[t]
\centering
\includegraphics[width=0.9\columnwidth]{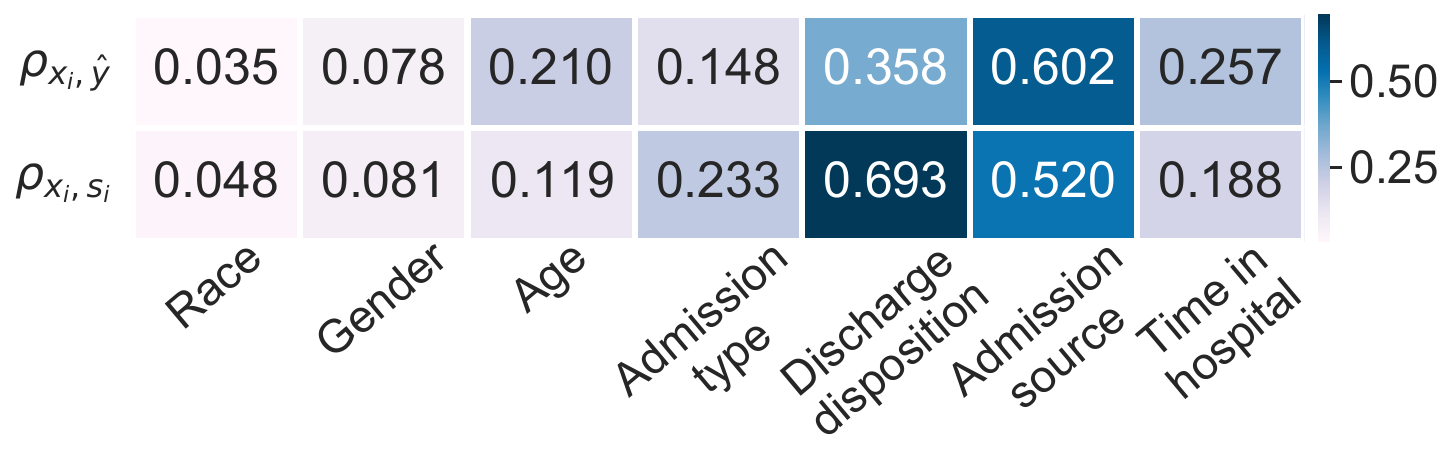}
\vspace{-4mm}
\caption{The Pearson correlation coefficient $\rho$ between the first seven features of the Diabetes dataset~\cite{diabetes} and the model output $\hat{y}$ as well as the corresponding explanations $\boldsymbol{s}$. NN and Google Cloud~\cite{Google} are  used as the testing model and platform.}
\label{fig-attack2-motivation}
\vspace{-3mm}
\end{figure}

\subsection{Adversary 2: Feature Inference Attack with Data Independence}\label{subsec-attack-2}
\subsubsection{Motivation.} Suppose the adversary has no auxiliary datasets $\mathcal{X}_{aux}$, it will be impossible
for the adversary to learn an attack model as discussed in Section~\ref{subsec-attack-1} because the distributions of the target private features are unknown. 
A naive way to restore $\boldsymbol{x}$ from $\boldsymbol{s}$ is that the adversary generates a random dataset $\mathcal{X}_{rand}$, sends the samples in this dataset to the MLaaS platform separately and obtains the corresponding explanations $\mathcal{S}_{rand}$. 
Provided that the size of the random dataset is large enough, there will be some $\boldsymbol{s}_{rand}\in \mathcal{S}_{rand}$ such that $||\boldsymbol{s}_{rand}-\boldsymbol{s}||_2 \leq \xi$ where $\xi$ is a small threshold. The adversary can use the $\boldsymbol{x}_{rand}$ associated with $\boldsymbol{s}_{rand}$ as the estimation of $\boldsymbol{x}$. 
But there are two problems in this method. 
First is that to improve the estimating accuracy, the adversary needs to randomly draw a large number of samples, e.g., $|\mathcal{X}_{rand}|>10000$, from the possible feature space and send them to the MLaaS platform, which can be costly on the pay-per-query service and increase the risk of being detected and blocked by the service provider. 
Second is that although $||\boldsymbol{s}_{rand}-\boldsymbol{s}||_2 \leq \xi$ holds, the corresponding $\boldsymbol{x}_{rand}$ can be a false positive case in which $||\boldsymbol{x}_{rand}-\boldsymbol{x}||_2$ is large. 
%

In the experiments, we observe that the level of correlations between a feature $x_i$ and the model output $\hat{y}$ is closely consistent with the correlations between $x_i$ and its explanation $s_i$. Fig.~\ref{fig-attack2-motivation} shows an example. 
First, we train a neural network on a real-world Diabetes dataset~\cite{diabetes} and deploy it to Google Cloud~\cite{Google}. Then, we send queries of $\boldsymbol{x}=\{x_i\}^n_{i=1}$ to the deployed model and obtain the model outputs $\hat{y}$ and explanations $\boldsymbol{s}=\{s_i\}^n_{i=1}$. We compute the Pearson correlation coefficients\footnote{\label{footnote-pearson}
{
$\rho(x, y)=\frac{\text{cov}(x, y)}{\sigma_x \sigma_y}$, where $\text{cov}(x, y)$ denotes the covariance between $x$ and $y$~\cite{Pearson}.
}
} between the first seven features and $\hat{y}$ as well as $\boldsymbol{s}$ and show them in Fig.~\ref{fig-attack2-motivation}, from which we observe that the values of $\rho_{x_i,\hat{y}}$ and $\rho_{x_i,s_i}$ are similar for all $x_i$.
Note that the Pearson correlation coefficient measures the linear correlation between variables~\cite{Pearson}, which inspires that to reduce the number of random queries and improve the estimation accuracies, we can exploit the  local linearity of the decision boundaries of the black-box model $f$ \textit{w.r.t.} some important features $x_i$. Then by interpolation based on a small random dataset $\mathcal{X}_{rand}$, we can accurately reconstruct the values of these features from the corresponding Shapley values $\boldsymbol{s}$.

\begin{algorithm}[!tb]
\begin{small}
\caption{The Attack with an Auxiliary Dataset}
\label{alg-attack-1}
\KwInput{the black-box model $f$, an auxiliary dataset $\mathcal{X}_{aux}$, a learning rate $\alpha$, the target Shapley values $\boldsymbol{s}$}
\KwOutput{The reconstructed private input $\hat{\boldsymbol{x}}$}
$\mathcal{S}_{aux}\gets \phi(\mathcal{X}_{aux}; f)$ \tcp*{Send queries to the ML platform}
$\boldsymbol{\theta}_{\psi} \leftarrow \mathcal{N}(0,1)$ \tcp*{Initialize the attack model}
\For{each epoch} {
    \For{each batch}{
        $loss \gets 0$ \;
        $B \gets$ randomly select a batch of samples \;
        \For{$j \in \{1, \cdots, |B|\}$}{
            $(\hat{\boldsymbol{x}}_{aux})^j \leftarrow {\psi}( (\boldsymbol{s}_{aux})^j; \boldsymbol{\theta}_{\psi})$ \;

            $loss$ += $L((\hat{\boldsymbol{x}}_{aux})^j, ({\boldsymbol{x}}_{aux})^j)$ \;
        }
        $\boldsymbol{\theta}_{\psi} \leftarrow \boldsymbol{\theta}_{\psi} - \alpha \cdot \bigtriangledown_{\boldsymbol{\theta}_{\psi}} loss$ \tcp*{Update the attack model}
    }
}
$\hat{\boldsymbol{x}}\gets \psi(\boldsymbol{s};\theta_{\psi})$ \;
\Return $\hat{\boldsymbol{x}}$\;
\end{small}
\end{algorithm}

\subsubsection{The Attack with A Random Dataset}\label{subsubsec-attack2}
In this attack, because the adversary has no information on the data distribution and feature interactions, different features in the random dataset $\mathcal{X}_{rand}$ have to be drawn independently, i.e., the features are independent of each other in $\mathcal{X}_{rand}$, which means that the features of $\boldsymbol{x}$ need to be restored separately from $\boldsymbol{s}$.
To achieve this task,
we introduce a generalized additive model (GAM)~\cite{hastie2017generalized} to approximate the black-box model $f$:
\begin{equation}\label{eq-gam}
  f(\boldsymbol{x})=f_1(x_1)+f_2(x_2)+\cdots + f_n(x_n) + \epsilon_f,
\end{equation}
where $\epsilon_f$ denotes the error caused by GAM approximation, and $f_i$ denotes an unknown smooth univariate function.
Note that GAM can theoretically minimize the Kullback-Leibler distance to the original function~\cite{hastie2017generalized} and has been used for analyzing the data flow in neural networks~\cite{agarwal2021neural, yang2021gami} and random forests~\cite{gregorutti2017correlation}.
Replace Eq.~\ref{eq-gam} into Eq.~\ref{eq-shap}, we have
\begin{equation}\label{eq-gam-shap}
  s_i=f_i(x_i)-f_i(x^0_i)+\epsilon_s,
\end{equation}
where $f_i(x^0_i)$ is a constant, and $\epsilon_s$ denotes the sampling error.
If $s_i$ is linearly correlated with $x_i$ (i.e., $f_i$ is monotonic around $x_i$), we can first find a set of $\{(x_{rand})^{j}_i\}^k_{j=1}$ from the random dataset $\mathcal{X}_{rand}$ with $|(s_{rand})^j_i-s_i|\leq \xi$ then estimate $x_i$ by $\hat{x}_i=\frac{1}{k}\sum^{k}_{j=1}(x_{rand})^{j}_i$.

\vspace{0.5mm}\noindent
\textbf{Bounding the estimation error.} Now we theoretically analyze the estimation error of the interpolation-based attack method. Since the features are reconstructed independently, it suffices to focus on only one feature in the target Shapley values $\boldsymbol{s}$, say $s_i$. Let $x_i$ be the private feature associated with $s_i$, $X$ be a random variable whose  Shapley value ${s}_X$ is \textit{close enough} to $s_i$, i.e., $|{s}_X-s_i|\leq \xi$, where $\xi$ is a small threshold. 
Let $\{\tilde{x}^j\}^k_{j=1}$ be $k$ independent sampling points of $X$, and $\hat{x}_i=\frac{1}{k}\sum_j \tilde{x}^j$ be the empirical mean of these $k$ points. Now by Chebyshev's inequality~\cite{Chebyshev}, we have
\begin{equation}\label{eq-chebyshev}
  \Pr (|x_i-\mu| \leq u\sigma)\geq 1 - \frac{1}{{u}^2},
\end{equation}
where $u$ denotes any real number with $u>1$, $\mu$ and $\sigma$ denotes the expectation and standard deviation of $X$, respectively.
Because we use an empirical mean $\hat{x}_i$ of $X$ instead of its expectation $\mu$ to estimate $x_i$, we need to further bound the deviation between $\hat{x}_i$ and $\mu$ via Hoeffding's inequality~\cite{Hoeffding}:
\begin{equation}\label{eq-hoeffding}
  \Pr (|\hat{x}_i - \mu | \leq w)\geq 1 -  2\exp \left( - \frac{2w^2 k}{(b-a)^2} \right),
\end{equation}
where $w$ denotes any real number with $w>0$, $[a, b]$ denotes the range of the $k$ random points, i.e., $a\leq\tilde{x}^j\leq b$ for all $j$. 
By combining Eq.~\ref{eq-hoeffding} with Eq.~\ref{eq-chebyshev}, we have
\begin{equation}\label{eq-error-bound}
  \Pr ( |x_i-\hat{x}_i|\leq u\sigma + w   )\geq 1 - \frac{1}{u^2} - 2\exp \biggl(\ln (u^2-1) - \frac{2w^2 k}{(b-a)^2} \biggr) \bigg/ u^2.
\end{equation}
Note that $\sigma < \frac{b-a}{2}$. The result of Eq.~\ref{eq-error-bound} shows that when the range $[a, b]$ of the sampling points $\{\tilde{x}^j\}^k_{j=1}$ becomes smaller, the estimation $\hat{x}_i$ becomes more accurate with a larger probability. For example, let $u=2$, $w=0.1$ and $k=30$. If $b-a=0.1$, we can reconstruct $x_i$ with an error less than 0.2 in a probability of larger than $75\%$. And if $b-a=0.5$, the estimation error becomes $0.6$ in a probability of $61\%$.

\begin{figure}[t]
\centering
\includegraphics[width=0.99\columnwidth]{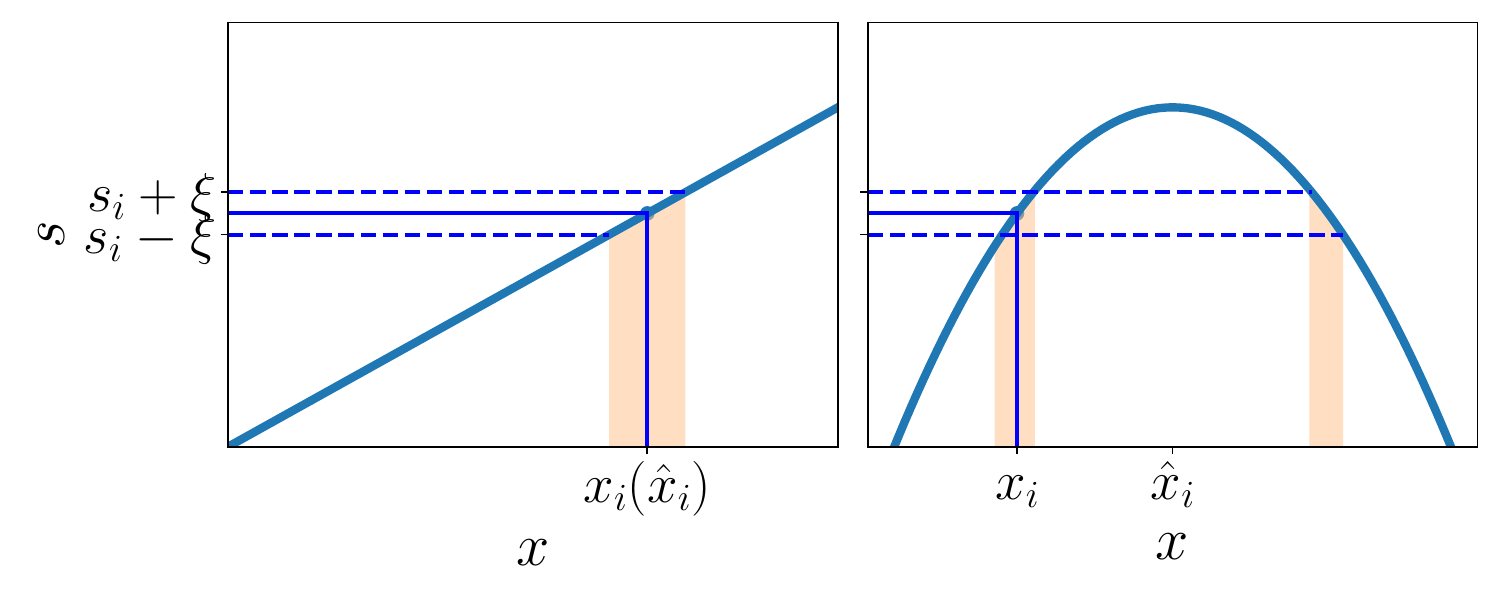}
\vspace{-4mm}
\caption{Examples of correlations between private features and the corresponding Shapley values.}
\label{fig-attack2-egg}
\vspace{-3mm}
\end{figure}

We explain the intuition behind the error bound via Fig.~\ref{fig-attack2-egg}, which depicts the possible correlations between $x_i$ and $s_i$.
The observation is that for important features, a slight change in their values will lead to a great difference in the model predictions and explanations (see the example in Fig.~\ref{fig-attack2-motivation}), which means that in general, these features are linearly correlated with their Shapley values, e.g., the left figure of Fig.~\ref{fig-attack2-egg}.
In this case, the range of estimations $[a, b]$ tends to be small, leading to an accurate reconstruction of these features. On the other hand, the less important features have little impact on the model predictions and explanations. 
This case corresponds to the right figure of Fig.~\ref{fig-attack2-egg}, i.e., the feature values scattered in a large range can produce a similar Shapley value $s_i$. And accordingly, $\sigma_{s_i} \leq \sigma_{\epsilon}$  (Eq.~\ref{eq-mutual-information-duction-4}) holds, making the estimation $\hat{x}_i$ far from the ground truth $x_i$ as the information of $x_i$ contained in $s_i$ is overridden by the approximation noise.

\begin{algorithm}[!tb]
\begin{small}
\caption{The Attack with Data Independence}
\label{alg-attack-2}
\KwInput{the black-box model $f$, the threshold for sampling error $\xi$, the minimum number of interpolation samples $m_{C}$, the threshold for the range of estimation values $\tau$, the target Shapley values $\boldsymbol{s}$}
\KwOutput{The reconstructed private input $\hat{\boldsymbol{x}}$}
$\hat{\boldsymbol{x}}\gets \emptyset$\;
$\mathcal{X}_{rand}\gets$ Random Samples \;   \label{alg2-line-generateX}
$\mathcal{S}_{rand}\gets \phi(\mathcal{X}_{rand}; f)$ \tcp*{Send queries to ML platform}  \label{alg2-line-query}
$m\gets |\mathcal{X}_{rand}|$ \;
$n\gets |\boldsymbol{s}|$ \;
\For(\tcp*[f]{For each feature}){$i=1,2,\dots,n$}{   \label{alg2-line-attack-start}
    $s_{t}\gets {s}_i$\;
    $D \gets \emptyset$             \tcp*{The distance set from $s_t$}
     \For{$j=1,2,\dots,m$}{ \label{alg2-line-dist-start}
        $dist \gets ||s_{t} - ({s}_{rand})^{j}_{i} ||_2$\;
        $\tilde{x} \gets ({x}_{rand})^{j}_{i}$\;
        $D \gets D \cup (dist, \tilde{x})$\;  \label{alg2-line-dist-end}
    }
    Sort $D$ on $dist$ in an ascending order\;  \label{alg2-line-sort}
     $C \gets \emptyset$ \tcp*{The candidate set of $\hat{x}_i$}
    \For{$j=1,2,\dots,m$}{
        $(dist, \tilde{x})\gets D_j$\;
        \If(\tcp*[f]{The Shapley error}){$|C|<m_{C}$ or $dist<\xi$}{   \label{alg2-line-shapley-error}
            $C \gets C \cup \tilde{x}$\;
        }
    }
    \If(\tcp*[f]{The estimation error}){$\max C - \min C > \tau $}{   \label{alg2-line-estimation-error}
        $\hat{x}_i \gets\bot $\;
    }\Else{
        $\hat{x}_i \gets \frac{1}{|C|}\sum C$\;
    }
    $\hat{\boldsymbol{x}}\gets\hat{\boldsymbol{x}} \cup \hat{x}_i$\;  \label{alg2-line-attack-end}
}
\Return $\hat{\boldsymbol{x}}$\;
\end{small}
\end{algorithm}

Algorithm~\ref{alg-attack-2} shows the attack details of adversary 2 without any background knowledge. First, we generate a random dataset $\mathcal{X}_{rand}$ (line~\ref{alg2-line-generateX}) and send it to the MLaaS platform to obtain an explanation set $\mathcal{S}_{rand}$ (line~\ref{alg2-line-query}). Then, we try to reconstruct the private features $\boldsymbol{x}$ associated with the target Shapley values $\boldsymbol{s}$ one by one (line~\ref{alg2-line-attack-start}-\ref{alg2-line-attack-end}).
For a target feature $x_i$, first, we compute the distances between its $\boldsymbol{s}_i$ and the Shapley values of the $i$-th features in the random samples (line~\ref{alg2-line-dist-start}-\ref{alg2-line-dist-end}).
After that, we sort the random features based on the Shapley value distances (line~\ref{alg2-line-sort}) and accordingly choose the candidate estimations of $x_i$ whose Shapley value distances are less than a threshold $\xi$.
Note that $\xi$ ($>\epsilon_s$) indicates the approximation error of Shapley sampling values (see Eq.~\ref{eq-shap-sampling} and \ref{eq-gam-shap}) and is used to impose the estimation error bound in Eq.~\ref{eq-error-bound} on the real value of $x_i$.
We use an empirical threshold $m_{C}$ as the minimum size of the candidate set to reduce possible estimation bias. In the end, we only reconstruct those private features with value ranges of the candidate estimations less than a threshold $\tau$ (line~\ref{alg2-line-estimation-error}).
The complexity of Algorithm~\ref{alg-attack-2} is $O(nm\log m)$, where $n$ denotes the number of target features and $m$ denotes the size of the random dataset, i.e., $|\mathcal{X}_{rand}|$.
It is worth noting that although the accuracy of estimation is limited by the sampling size $m$, in the experiments, we show that $m=100$ suffices to reconstruct more than $30\%$ features with high accuracies from the complicated models (e.g., NN).

\subsection{Attack Generalization}\label{subsec-attack-generalization}
{It is worth noting that our attacks are designed based on the inherent limitations of Shapley values. Specifically, we can simplify the computation of Shapley values into three steps: first use an affine transformation to generate a feature replacement dataset from $x$ and $x^0$, then feed the dataset into the target model, and finally use another affine transformation on the model outputs to obtain Shapley values.
Because the mapping from private features to Shapley values is fixed, we can then propose the inverse mapping attack (attack 1) and the GAM transformation-based attack (attack 2) to reconstruct the inputs from their explanations.  }

{On the contrary, 
other popular explanation methods, e.g., LIME~\cite{ribeiro2016LIME} and DeepLIFT~\cite{shrikumar2017DeepLIFT}, focus on approximating the linear decision boundary around the target features via a set of heuristic rules, where the mappings between features and explanations among different samples are unstable and various~\cite{ancona2017towards}, leading to little convergence of attack 1 and invalidating the GAM transformation and error bounds of attack 2. For justification, we compared the performance of attack 1 on Shapley values, LIME, and DeepLIFT tested on the Google Cloud platform~\cite{Google}, where the testing model and dataset are NN and Adult~\cite{UCI}. 
Note that attack 2 is not applicable to other explanation methods because the GAM transformation is based on the computational formula of Shapley values. 
For attack 1, the averaging $\ell_1$ error of feature reconstruction tested on Shapley values is 0.0768, which is far better than 0.1878 on LIME, 0.1527 on DeepLIFT, and 0.1933 on the empirical random guess baseline (see Section~\ref{subsec-exp-setting}).
The conclusion is that the proposed attacks are targeted on Shapley values, and specific optimizations are needed to apply them to other heuristic explanation methods. We leave it as our feature work.}

\section{Experiments}\label{sec-exp}

\subsection{Experimental Setting}\label{subsec-exp-setting}
The proposed algorithms are implemented in Python with \textit{PyTorch}~\footnote{https://pytorch.org/}, and all attack experiments are performed on a server equipped with Intel (R) Xeon (R) Gold 6240 CPU @ 2.60GHz$\times$72 and 376GB RAM, running Ubuntu 18.04 LTS.

\vspace{0.5mm}\noindent
\textbf{Explanation Platforms.}
To validate the effectiveness of the proposed attacks on  real-world applications, we conduct experiments on three leading ML platforms: \textit{Google} Cloud AI platform~\cite{Google}, \textit{Microsoft} Azure Machine Learning~\cite{Microsoft} and \textit{IBM} Research Trusted AI~\cite{IBM}. For the experiments on Google and Microsoft platforms, we first train models and deploy them in the cloud, then send prediction and explanation queries to the deployed models to obtain the corresponding explanation vectors. 
The experiments on the IBM platform are performed locally because a python package called AI Explainability 360~\cite{IBM} is available.
In addition, to test the attack performance \textit{w.r.t.} different sampling errors, we implement a \textit{Vanilla} Shapley sampling method~\cite{vstrumbelj2014explaining}, because the explanatory API provided by Google only supports up to 50 permutations for sampling computation (e.g., $\upsilon\leq 50$ in Section~\ref{subsec-shapley-value}) while the APIs provided by Microsoft and IBM are based on SHAP~\cite{lundberg2017SHAP} which fails to provide a theoretical error bound for Shapley approximation. 
The experiments are performed on the Google platform by default unless otherwise specified.

\vspace{0.5mm}\noindent
\textbf{Datasets.} Six real-world datasets are used in the experiments: \textit{Adult}~\cite{UCI} for predicting whether one's income will exceed 50000 per year, 
%
\textit{Bank} Marketing~\cite{bank} for predicting whether a client will subscribe a deposit, 
\textit{Credit} Card Clients~\cite{credit} for predicting whether  a client is credible based on payment recordings, 
\textit{Diabetes} 130-US Hospitals~\cite{diabetes} for predicting whether a patient will be readmitted,
{\textit{IDA} 2016 Challenge~\cite{UCI} for predicting whether a component failure in a truck is related to the air pressure system,
and \textit{Insurance} Company Benchmark~\cite{insurance} for predicting whether a customer has a caravan insurance policy. }
%
%
In addition, we use three synthetic datasets generated via the \textit{sklearn}~\footnote{https://scikit-learn.org/stable/} library to evaluate the impact of correlations between input features and model outputs on the attack performance. The feature values in all datasets are normalized into $[0, 1]$ for ease of comparison~\cite{luo2021feature, luo2021fusion}. The details of these datasets are summarized in Tab.~\ref{tab-datasets}.
For each dataset $\mathcal{X}$, we randomly split it into three portions: a dataset $\mathcal{X}_{train}$ with $60\%$ samples for model training, a dataset $\mathcal{X}_{test}$ with $20\%$ samples for model testing and acting as the auxiliary dataset collected by the attacker, a dataset $\mathcal{X}_{val}$ with the rest samples for attack validation. 

\begin{table}[t!]
\caption{The Experimental Datasets.}\label{tab-datasets}
\centering
\vspace{-2mm}
\begin{tabular}{  l  c c c }
\toprule
Dataset & Instances & Classes & Features \\
\toprule
Adult & 48842 & 2 & 14\\
Bank marketing &	45211&	2&	20\\
Credit card	&30000&	2	&23\\
Diabetes 130-US hospitals	&68509&	3&	24\\
{IDA 2016 Challenge} & 76000 & 2 & 170 \\
{Insurance Company Benchmark} & 90000 & 2 & 85 \\
Synthetic dataset$\times$3 	& 100000 &	5&	12\\
\bottomrule
\end{tabular}
\vspace{-3mm}
\end{table}

\vspace{0.5mm}\noindent
\textbf{Black-Box Models.} Four types of models are used in the experiments: Neural Network (NN), Random Forest (RF),  Gradient Boosted Decision Trees (GBDT) and  Support-Vector machine (SVM) with the Radial Basis Function (RBF) kernel. The NN model consists of an input layer with $n$ dimensions, an output layer with $c$ dimensions, and two hidden layers with $2n$ neurons per layer. The number of trees and maximum depth for each tree are set to $\{100,5\}$ for RF and $\{100, 3\}$ for GBDT.
The models deployed on Google Cloud are trained with TensorFlow~\footnote{https://www.tensorflow.org/}, while the models on other platforms are trained with Pytorch (NN) and sklearn (RF, GBDT, SVM). The initial states of models and training hyper-parameters among different platforms are set to be the same for consistency.

\vspace{0.5mm}\noindent
\textbf{Attack Algorithms.} For adversary 1, we train a multi-layer perceptron as the attack model $\psi$, which consists of one input layer with $n$ dimensions, one output layer with $n$ dimensions, and one hidden layer with $4n$ dimensions. Sigmoid is used as the activation function of both the hidden and output layers.
The reason is that MLPs with Sigmoid activation can approximate any models with a desired accuracy~\cite{hornik1989multilayer} and are suitable for learning a pseudo inverse mapping of the black-box models with unknown structures. For adversary 2, we set the minimum number of the interpolation samples $m_{C}$ in Algorithm~\ref{alg-attack-2} to be 30, and the threshold $\tau$ for the range of estimation values to be 0.4. For the threshold of sampling error $\xi$, considering that the intensities of Shapley values in different datasets are also different, we determine $\xi$ based on the tight range $r=\max \mathcal{S} - \min \mathcal{S}$ of the available Shapley values $\mathcal{S}$ for different datasets and set it to be $\frac{r}{5}$ by default. Note that $\mathcal{S}$ can be generated from $\mathcal{X}_{aux}$ or $\mathcal{X}_{rand}$.

\vspace{0.5mm}\noindent
\textbf{Metrics.}
We use the Mean Absolute Error (MAE, i.e., $\ell_1$ loss) to measure the distances between the reconstructed private inputs $\hat{\boldsymbol{x}}$ and the ground truth inputs $\boldsymbol{x}$:
\begin{equation}\label{eq-l1}
  \ell_1 (\hat{\boldsymbol{x}}, \boldsymbol{x}) =\frac{1}{mn}\sum_{j=1}^{m}\sum_{i=1}^{n}|\hat{x}^j_i-x^j_i|,
\end{equation}
where $m$ is the number of samples in the validation dataset $\mathcal{X}_{val}$ and $n$ is the number of features.
Besides, we use the success rate (SR) to evaluate the percentage of features that can be successfully reconstructed by attack 2:
\begin{equation}\label{eq-sr}
  \text{SR}=\frac{|\hat{\mathcal{X}}_{val} \neq \bot|}{mn},
\end{equation}
where $|\hat{\mathcal{X}}_{val} \neq \bot|$ represents the total number of features that are successfully restored in $\mathcal{X}_{val}$.
For each experiment setting, we first randomly select 10 independent samples from $\mathcal{X}_{train}$ as the reference samples $\boldsymbol{x}^0$ (see Section~\ref{subsec-shapley-value}), then accordingly conduct 10 independent experiments and report the averaging results.

\vspace{0.5mm}\noindent
\textbf{Baselines.}
For adversary 1, since the attacker owns an auxiliary dataset that shares the same underlying distribution $P_{\mathcal{X}}$ of the private inputs $\boldsymbol{x}$, we use a random guess baseline based on the empirical distribution (called RG-E) that estimates the values of $\boldsymbol{x}$ by randomly sampling a vector $\tilde{\boldsymbol{x}}\sim P_{\mathcal{X}}$. For adversary 2, since the attacker does not know the distribution of the private inputs, we use two random guess baselines to estimate the values of $\boldsymbol{x}$: a Uniform distribution $\boldsymbol{U}(0,1)$ (called RG-U) and a Gaussian distribution $\boldsymbol{N}(0.5,0.25^2)$ (called RG-N). 
{
Note that RG-N can generate random values falling in $[0, 1]$ with a probability of at least $95\%$~\cite{luo2021feature}. }

\begin{figure*}[t]
\centering
\begin{small}
\begin{tabular}{cccc}
\multicolumn{4}{c}{\hspace{0mm} \includegraphics[height=4mm]{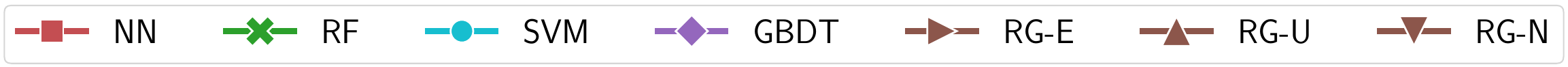}}
\vspace{-2mm}  \\
\hspace{-3mm}
\subfigure[{Credit}]{\includegraphics[width=0.25\textwidth]{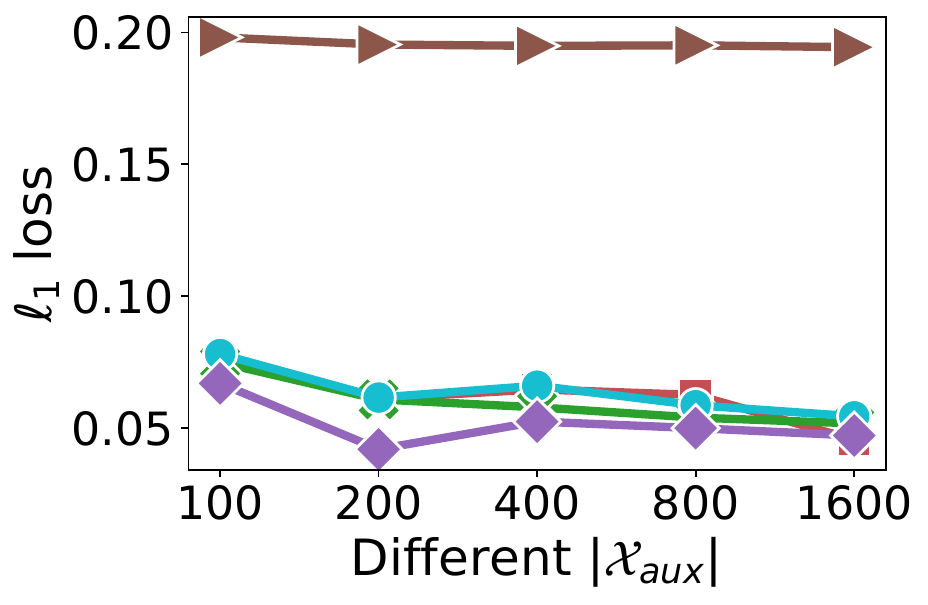}\label{subfig-sizes-attack1-start}}
&
\hspace{-3mm}
\subfigure[{Diabetes}]{\includegraphics[width=0.25\linewidth]{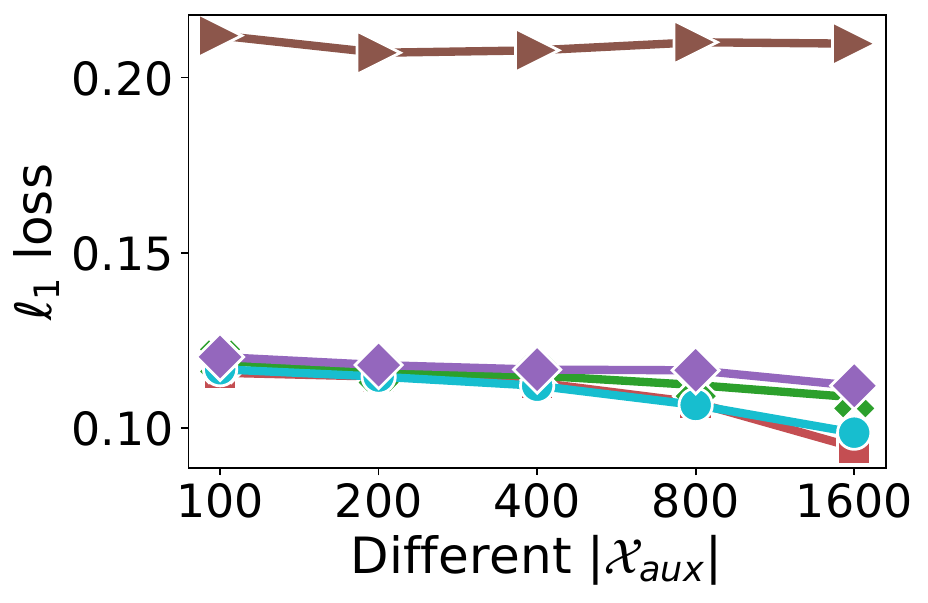}}
&
\hspace{-3mm}
\subfigure[{{IDA}}]{\includegraphics[width=0.25\linewidth]{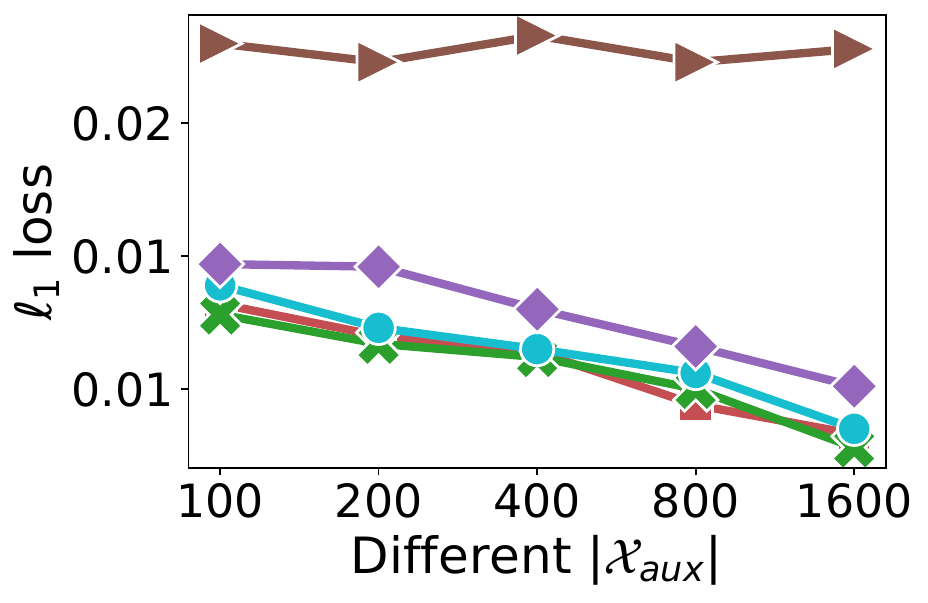}\label{subfig-size-1-ida}}
&
\hspace{-3mm}
\subfigure[{{Insurance}}]{\includegraphics[width=0.25\linewidth]{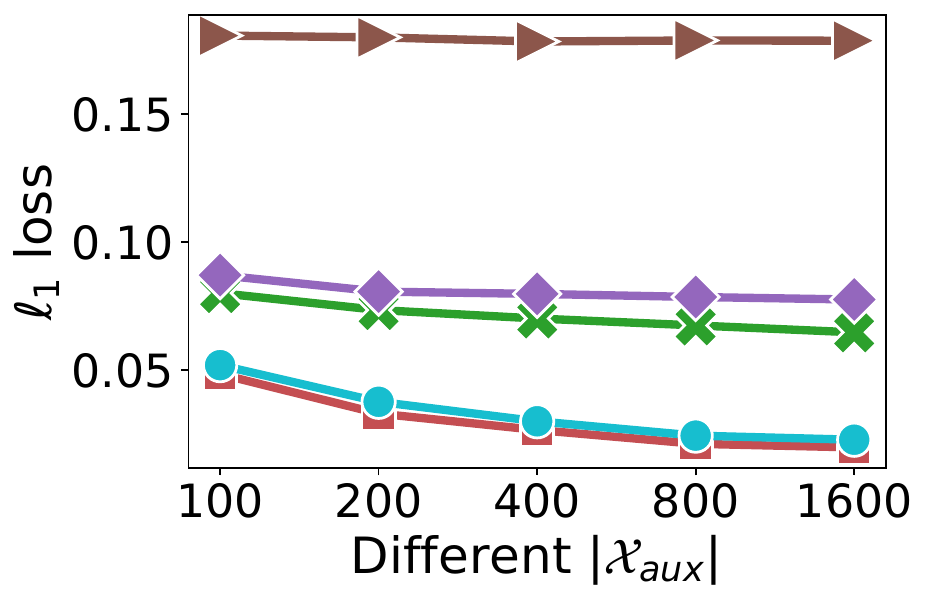}\label{subfig-sizes-attack1-end}} 
\vspace{-2mm}\\
\hspace{-3mm}
\subfigure[Credit]{\includegraphics[width=0.25\textwidth]{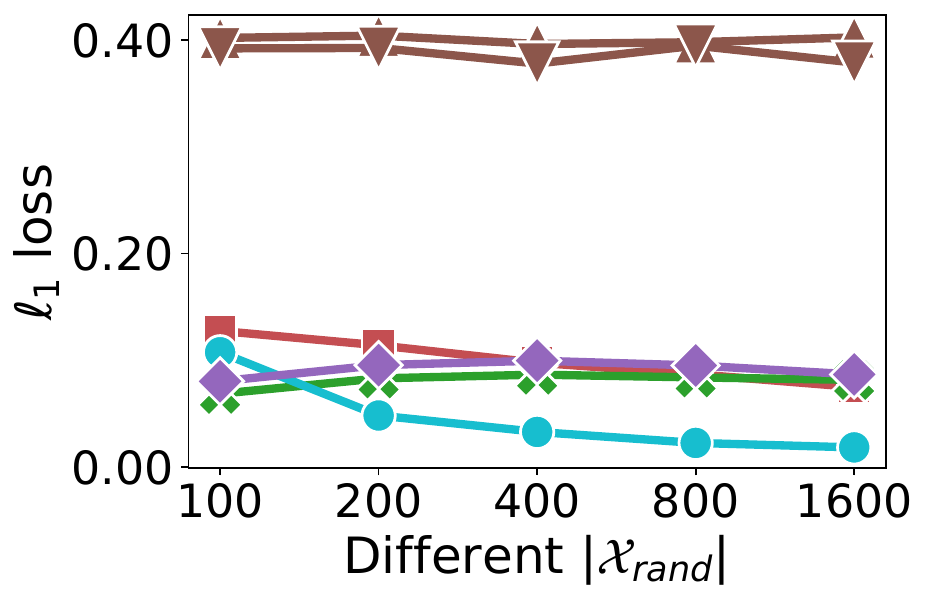}\label{subfig-sizes-attack2-start}}
&
\hspace{-3mm}
\subfigure[Diabetes]{\includegraphics[width=0.25\textwidth]{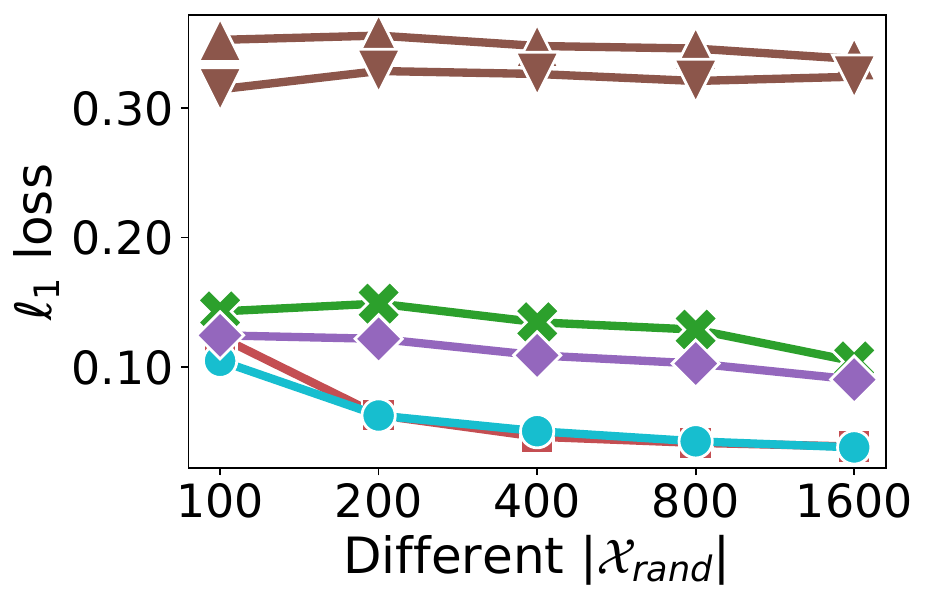}\label{subfig-size-2-diabetes}}
&
\hspace{-3mm}
\subfigure[{IDA}]{\includegraphics[width=0.25\textwidth]{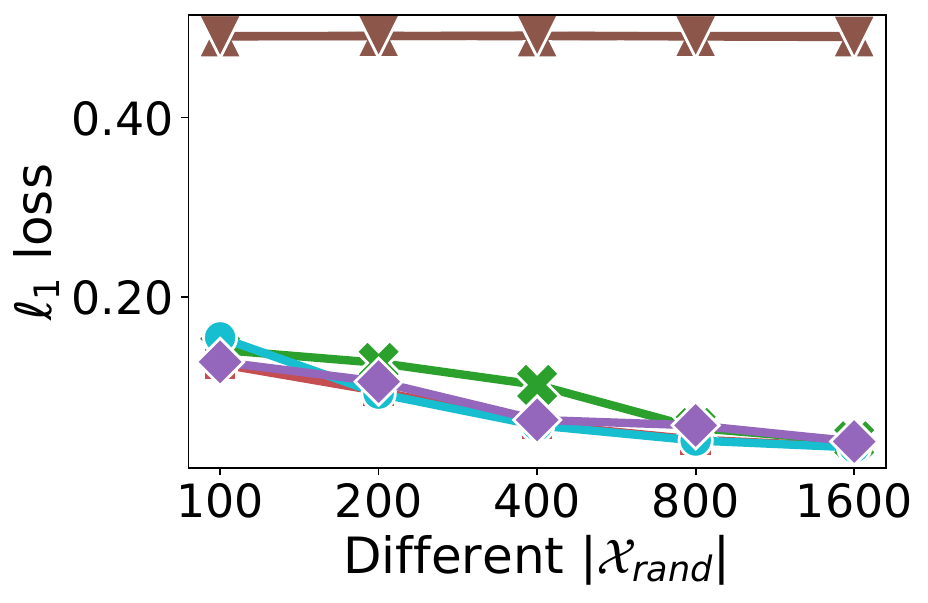}\label{subfig-size-2-ida}}
&
\hspace{-3mm}
\subfigure[{Insurance}]{\includegraphics[width=0.25\textwidth]{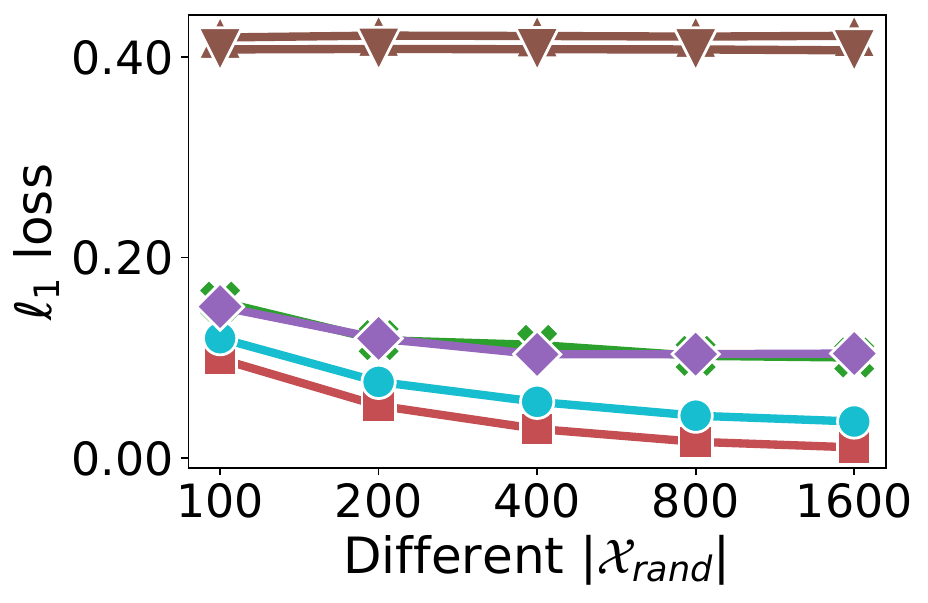}\label{subfig-sizes-attack2-end}}
\end{tabular}
\vspace{-5mm}
\caption{(a)-(d): the performance of attack 1 \textit{w.r.t.} different sizes of auxiliary datasets; (e)-(h): the performance of attack 2 \textit{w.r.t.} different sizes of random datasets.}
\label{fig-loss-sizes}
\end{small}
\vspace{-3mm}
\end{figure*}

\begin{figure*}[htbp]
\centering
\begin{small}
\begin{tabular}{cccc}
\multicolumn{4}{c}{\hspace{0mm} \includegraphics[height=4mm]{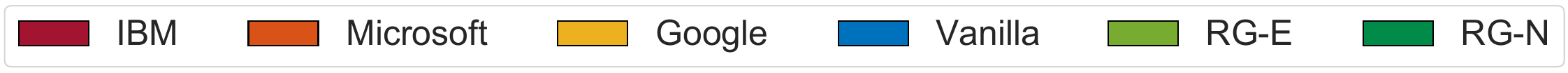}}
\vspace{-2mm}  \\
\hspace{-3mm}
\subfigure[Adult]{\includegraphics[width=0.25\textwidth]{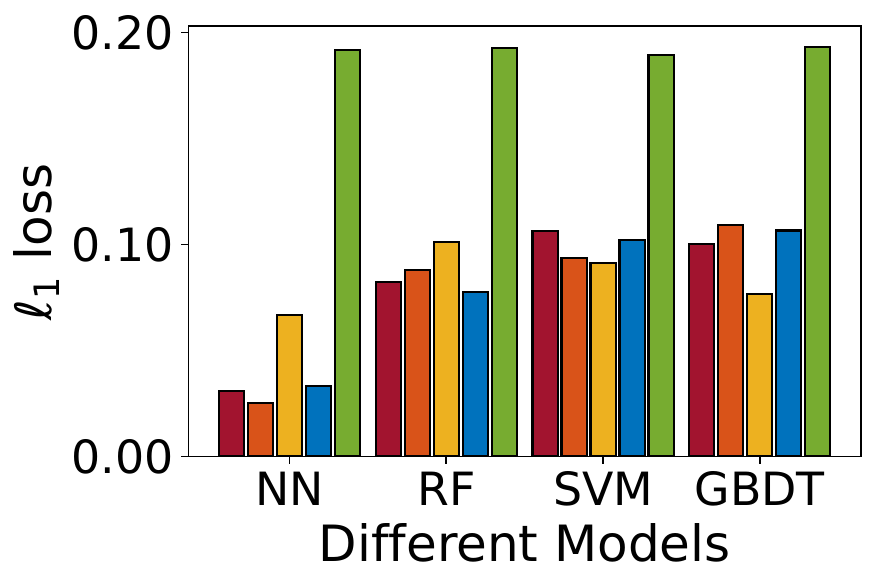}\label{subfig-platform-attack1-start}}
&
\hspace{-3mm}
\subfigure[Bank]{\includegraphics[width=0.25\textwidth]{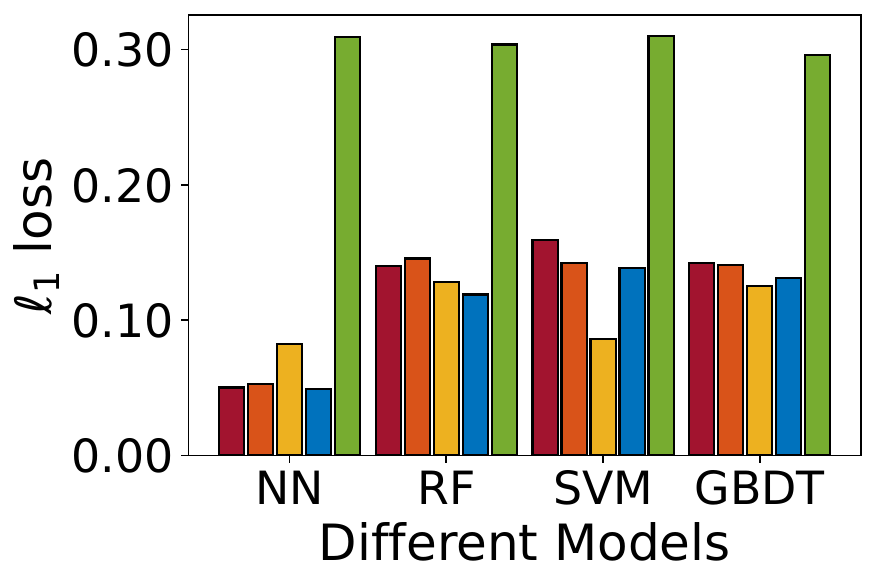}}
&
\hspace{-3mm}
\subfigure[Credit]{\includegraphics[width=0.25\textwidth]{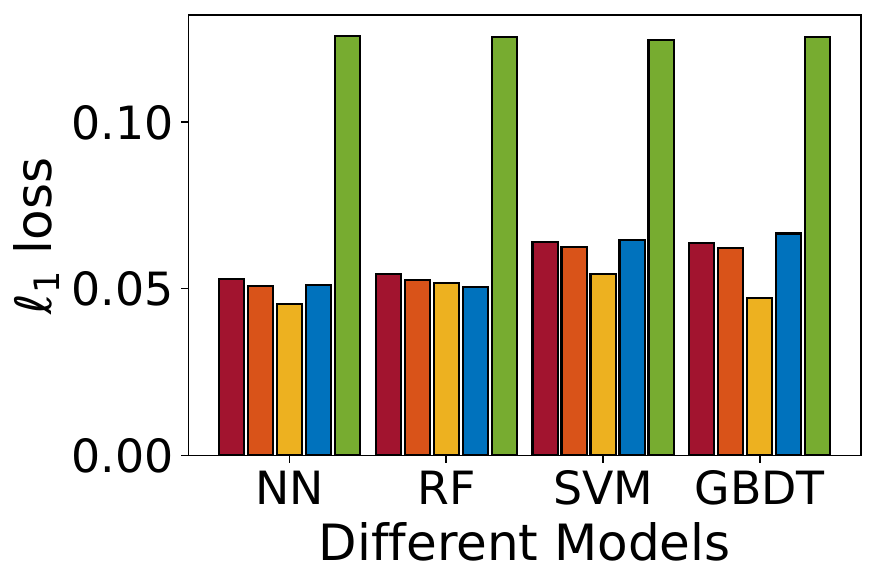}}
&
\hspace{-3mm}
\subfigure[Diabetes]{\includegraphics[width=0.25\textwidth]{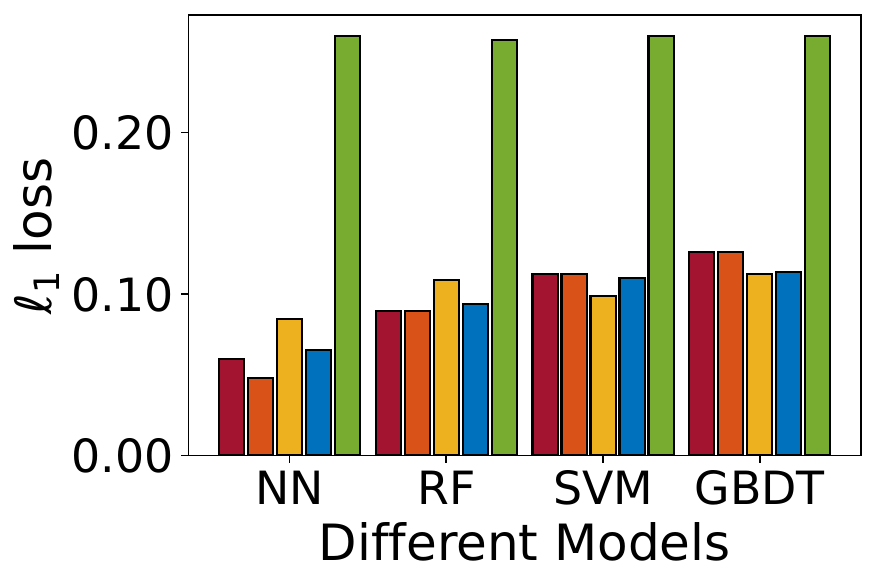}\label{subfig-platform-attack1-end}}
\vspace{-2mm}  \\
\hspace{-3mm}
\subfigure[Adult]{\includegraphics[width=0.25\textwidth]{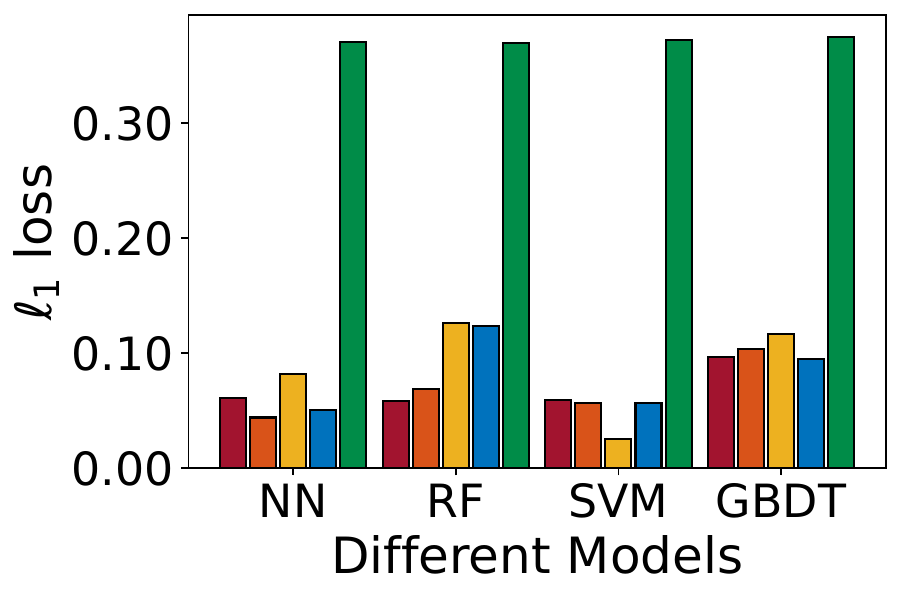}\label{subfig-platform-attack2-start}}
&
\hspace{-3mm}
\subfigure[Bank]{\includegraphics[width=0.25\textwidth]{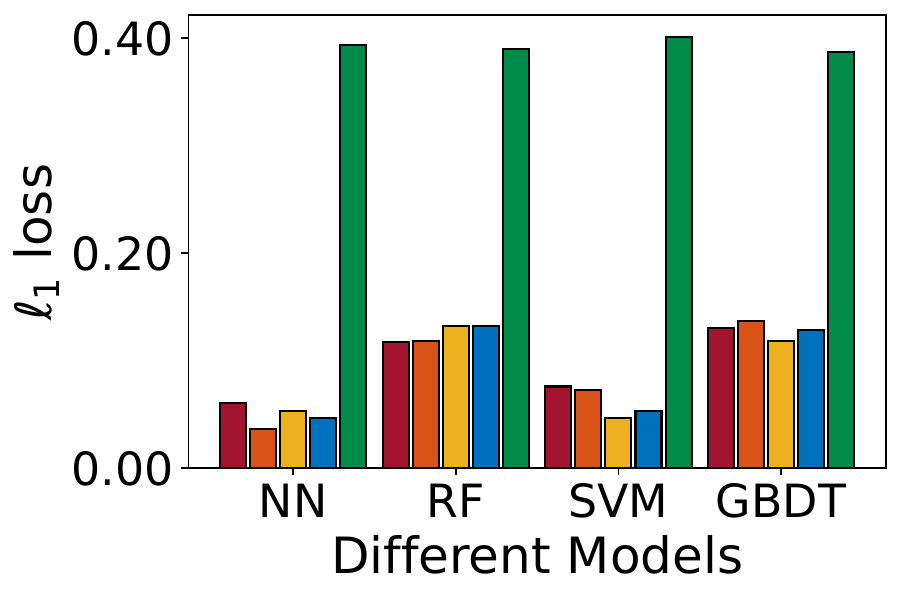}}
&
\hspace{-3mm}
\subfigure[Credit]{\includegraphics[width=0.25\textwidth]{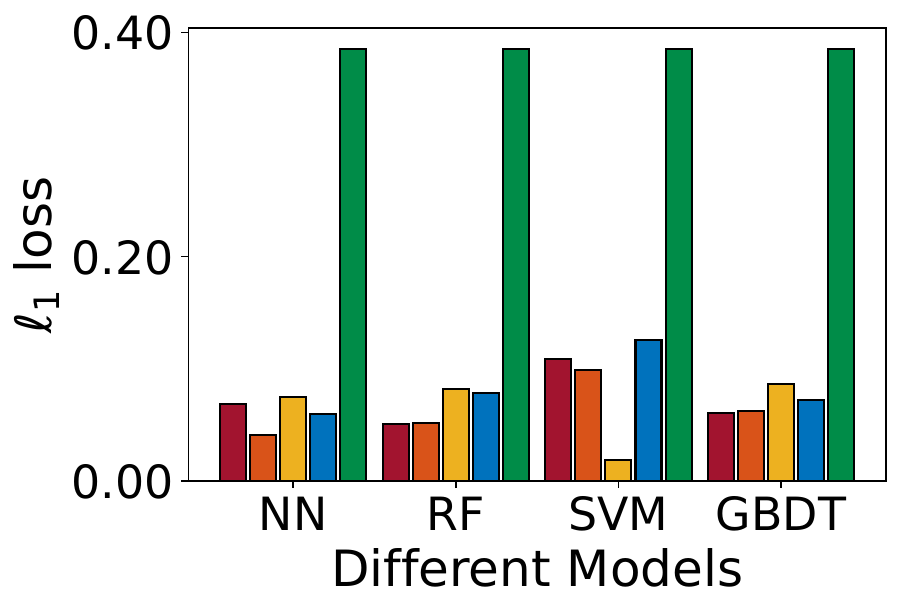}}
&
\hspace{-3mm}
\subfigure[Diabetes]{\includegraphics[width=0.25\textwidth]{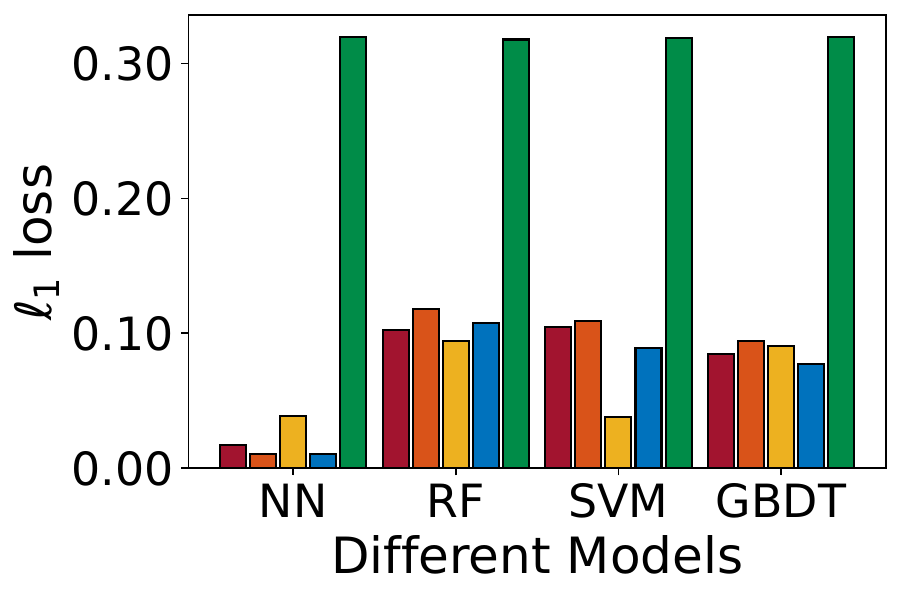}\label{subfig-platform-attack2-end}}
\end{tabular}
\vspace{-5mm}
\caption{The performance of the proposed attacks tested on different platforms: (a)-(d) attack 1, (d)-(e) attack 2.}
\label{fig-platforms}
\end{small}
\vspace{-3mm}
\end{figure*}

\subsection{Attack Performance \textit{w.r.t.} Different Numbers of Queries}\label{subsec-exp-sizes}
In both attacks, the adversary sends queries to the MLaaS platforms in a pay-per-query pattern. If the number of queries needed is large, the cost for the adversary could be unacceptable.
In this section, we evaluate the performance of attack 1 and 2 under different number of queries. The number of permutations used to compute Shapley sampling values is set to 50. 
We first vary the number of queries, i.e., the size of the auxiliary dataset $|\mathcal{X}_{aux}|$ or the random dataset $|\mathcal{X}_{rand}|$, in $\{100, 200, 400, 800, 1600\}$, then test the performance of attack 1 and 2, respectively. 
Fig.~\ref{subfig-sizes-attack1-start}-\ref{subfig-sizes-attack1-end} show the $\ell_1$ losses of attack 1. Fig.~\ref{subfig-sizes-attack2-start}-\ref{subfig-sizes-attack2-end} and Tab.~\ref{tb-success-sizes-attack2} show the $\ell_1$ losses and success rates of attack 2.
We have two observations from the results.

First, with the increase of queries, the estimation errors of adversary 1 and 2 slightly decrease (Fig.~\ref{subfig-sizes-attack1-start}-\ref{subfig-sizes-attack2-end}), and the success rates of adversary 2 generally increase (Tab.~\ref{tb-success-sizes-attack2}). The reason is straightforward: in adversary 1, the attack model $\psi$ can obtain a better generalization performance when the training dataset $\mathcal{X}_{aux}$ becomes larger; for adversary 2, a finer sampling of random points can help produce  smaller value ranges of the candidate estimations on the private features, thus reducing the estimation error bound (see Eq.~\ref{eq-error-bound}) and leading to better attack accuracies.
Meanwhile, the improvements in the attack accuracies are limited. For example, when varying the number of queries from 100 to 1600, the losses only decrease by 0.03 in adversary 1 and 0.05 in adversary 2 on the NN model of the Credit dataset. This means that a small number of queries (typically 100) suffice to initialize successful attacks.

Second, the proposed attacks perform worse on the ensemble models (RF and GBDT) than on NN and kernel SVM, e.g., Fig.~\ref{subfig-sizes-attack1-end}, \ref{subfig-size-2-diabetes} and \ref{subfig-sizes-attack2-end}. For adversary 1, the key idea is to learn the inverse mapping of the black-box models with a three-layer MLP. 
%
%
Each of the ensemble models used in our experiments consists of 100 decision trees. Compared to NN and kernel SVM, the decision boundaries of the ensemble models are more complicated and difficult to be accurately imitated by a simple neural network~\cite{biau2019neural, reinders2019neural}. Therefore, the generalization performances of the attack models trained on ensemble models become worse than on other models.
For adversary 2, the decision boundaries of RF and GBDT are more complicated than that of NN and SVM, leading to fewer local linearities between features and Shapley values, and thus larger estimation errors. As discussed in Section~\ref{subsec-preliminary-ml}, each tree in the ensemble models is typically trained on a set of randomly selected samples~\cite{sagi2018ensemble}. This method can prevent overfitting meanwhile introducing more randomness to the decision boundaries. The decision boundaries of NN models, on the other hand, are relatively smoother. In addition,  the linearities between the important features and model outputs can be effectively preserved in NN, because the activation functions intercept little information of these features.


\subsection{Attack Performance \textit{w.r.t.} Different MLaaS Platforms}
In this section, we evaluate the proposed attacks on different MLaaS explanation platforms. Because the methods for computing Shapley sampling values among different platforms are different, we implement a Vanilla Shapley sampling method based on \cite{vstrumbelj2014explaining} for comparison. The number of permutations sampled for computing Shapley values is set to 50 in both the Vanilla method and Google Cloud platform. Note that this parameter is not supported in the Microsoft Azure and IBM aix360 platforms because their explanation methods are based on SHAP~\cite{lundberg2017SHAP}, which computes Shapley values via a heuristic regression method instead of sampling permutations. The numbers of queries in both of the adversaries are set to 1000. The Shapley sampling error $\xi$ in adversary 2 (Algorithm~\ref{alg-attack-2}) is empirically set to $\frac{r}{5}$ among all platforms for comparison.
Fig.~\ref{subfig-platform-attack1-start}-\ref{subfig-platform-attack1-end} and \ref{subfig-platform-attack2-start}-\ref{subfig-platform-attack2-end} show the performance of adversary 1 and 2, respectively. Tab.~\ref{tb-success-platforms-attack2} shows the success rates of adversary 2 corresponding to Fig.~\ref{subfig-platform-attack2-start}-\ref{subfig-platform-attack2-end}.

From Fig.~\ref{subfig-platform-attack1-start}-\ref{subfig-platform-attack1-end} we observe that the attacks performed on Microsoft and IBM platforms achieve similar performance, which is expectable because these two platforms compute Shapley values via the same regression method. 
We also observe slight differences between the attacks performed on Vanilla and Google platforms although both of their implementations  are based on the same Shapley sampling method~\cite{vstrumbelj2014explaining}. 
Note that for efficiency, the maximum number of sampling permutations in Google Cloud is fixed to 50, which greatly reduces the computation costs from $O(2^n)$ to $O(50)$, meanwhile producing relatively large sampling error. As discussed in Section~\ref{subsec-attack-1}, a large error can override the information of private features in the Shapley values, leading to unstable reconstructions (see Eq.~\ref{eq-mutual-information-duction-4}).
Nevertheless, the proposed attacks can still reconstruct the private features with high accuracy.

\begin{table}[t!]
\center
\small
\caption{The success rates of attack 2 \textit{w.r.t.} different $|\mathcal{X}_{rand}|$.}\label{tb-success-sizes-attack2}
\vspace{-2mm}
\begin{tabular}{c|c|ccccc}
\hline
\multirow{2}{*}{Dataset} &\multirow{2}{*}{Model} & \multicolumn{5}{c}{Different $|\mathcal{X}_{rand}|$} \\
\cline{3-7}
  &   & 100  & 200 & 400 & 800 & 1600  \\
\hline
\hline
\multirow{4}{*}{Credit}     & NN & 0.5871& 	0.7220& 	0.7520& 	0.7928& 	0.7961\\
                            & RF & 0.4515& 	0.4976& 	0.5307& 	0.5833 &	0.5858\\
                            & SVM & 0.5575& 	0.7607& 	0.9404& 	0.9549& 	0.9552\\
                            & GBDT & 0.4680& 	0.6721& 	0.6833& 	0.6984& 	0.7486\\
\hline
\multirow{4}{*}{Diabetes}     & NN & 0.3911 &	0.6663 	&0.7509 	&0.7947 	&0.8262\\
                            & RF & 0.2798 	&0.3368 	&0.3830 	&0.4113 	&0.4469\\
                            & SVM & 0.3763 	&0.5429 	&0.6826 	&0.7241 	&0.6975\\
                            & GBDT & 0.3151 &	0.4176 	&0.4389 	&0.5043 	&0.5361 \\
\hline
\multirow{4}{*}{{IDA}}        & NN & 0.7835 & 	0.9896& 	0.9972& 	0.9972& 	0.9973  \\
                            & RF & 0.3114 &	0.3464 &	0.3763& 	0.4306& 	0.4857 \\
                            & SVM & 0.8965 &	0.9614& 	0.9671& 	0.9581& 	0.9638  \\
                            & GBDT & 0.2854 &	0.3171& 	0.3463& 	0.3515& 	0.3849 \\

\hline
\multirow{4}{*}{{Insurance}}  & NN & 0.9523& 	0.9684& 	0.9708& 	0.9688& 	0.9696\\
                            & RF & 0.3194& 	0.3339& 	0.3833& 	0.4399& 	0.4518 \\
                            & SVM & 0.6054& 	0.6764& 	0.7335& 	0.7529& 	0.7539\\
                            & GBDT & 0.3043& 	0.3072& 	0.3176& 	0.3268& 	0.3446\\
\hline
\end{tabular}
\vspace{-4mm}
\end{table}

From Tab.~\ref{tb-success-platforms-attack2}, we observe that the success rates of adversary 2 performed on IBM and Microsoft platforms are lower than the success rates on the Vanilla method and Google Cloud. Considering that SHAP~\cite{lundberg2017SHAP} computes Shapley values heuristically without providing a theoretical approximation bound, the real sampling errors computed by IBM and Microsoft could be large.  
Although the parameter $\xi$ in Algorithm~\ref{alg-attack-2} can guarantee that the real features fall into the estimation range $[a, b]$, the reconstructions on IBM and Microsoft could still fall in cases where the real Shapley sampling error is far larger than $\xi$. But in Tab.~\ref{tb-success-platforms-attack2}, we observe that the success rates of adversary 2 performed on IBM and Microsoft are at least $30\%$, which demonstrates the effectiveness of the proposed attacks.


\subsection{Attack Performance \textit{w.r.t.} Different Feature Importance}\label{subsec-exp-feature}
As discussed in Section~\ref{subsec-attack-2}, the important features can be more accurately reconstructed than the less important features. In this section, instead of averaging the reconstruction loss over all features, we give the attack loss per feature and dissect the connections between attack accuracies and feature importance.

\begin{table}[t!]
\center
\small
\caption{The success rates of attack 2 \textit{w.r.t.} different platforms.}\label{tb-success-platforms-attack2}
\vspace{-2mm}
\begin{tabular}{c|c|ccccc}
\hline
\multirow{2}{*}{Dataset} &\multirow{2}{*}{Model} & \multicolumn{4}{c}{Different Platforms} \\
\cline{3-6}
  &   & IBM  & Microsoft & Google & Vanilla   \\
\hline
\hline
\multirow{4}{*}{Adult}      & NN  & 0.9119  & 	0.9066  & 	0.9229  & 	0.9788  \\
                            & RF  &  0.5207  & 	0.5221  & 	0.5764  & 	0.6181\\
                            & SVM & 0.5209  & 	0.5192  & 	0.6964  & 	0.6460  \\
                            & GBDT &  0.4086  & 	0.4054  & 	0.4337  & 	0.6426 \\

\hline
\multirow{4}{*}{Bank}         & NN  &  0.8808  & 	0.9301  & 	0.9803  & 	0.9807\\
                            & RF    & 0.4133  & 	0.4241  & 	0.4857  & 	0.4954 \\
                            & SVM   &  0.4699  & 	0.4561  & 	0.6467  & 	0.6396\\
                            & GBDT  & 0.3404  & 	0.3384  & 	0.5993  & 	0.4774 \\
\hline
\multirow{4}{*}{Credit}     & NN  & 0.6405  & 	0.6913  & 	0.7961  & 	0.8784\\
                            & RF &  0.3913  & 	0.4018  & 	0.5858  & 	0.4858 \\
                            & SVM & 0.4642  & 	0.4759  & 	0.6552  & 	0.6295  \\
                            & GBDT & 0.3787  & 	0.3389  & 	0.4486  & 	0.4595  \\
\hline
\multirow{4}{*}{Diabetes}   & NN    & 0.8863  & 	0.9999  & 	0.8262  & 	1.0000 \\
                            & RF   & 0.4733  & 	0.4843  & 	0.5469  & 	0.5598 \\
                            & SVM &  0.5079  & 	0.5349  & 	0.6975  & 	0.6027  \\
                            & GBDT & 0.4111  & 	0.4011  & 	0.5361  & 	0.5058  \\
\hline
\end{tabular}
\vspace{-4mm}
\end{table}

Three synthesis datasets with 12 features are used for the experiments (see Tab.~\ref{tab-datasets} for the details). To generate the features of different importance, we first randomly draw five point clusters around five vertices of a three-dimension cube and label each cluster with a unique class. The three features of these random three-dimensional points are called the key features. After that, we generate $n_r$ redundant features by randomly and linearly combining the key features. The rest $12-3-n_r$ features are generated from random noises. The key features and redundant features are called important features.
We vary the percentages of important features in $\{25\%, 50\%,75\%\}$ and accordingly generate three synthesis datasets. In addition, two real-world datasets, Credit and Diabetes, are also used for justification.
%
%
The importance of feature $x_i$ is defined as the Mean Absolute Correlation Coefficients (MACC) between $x_i$ and model outputs $\boldsymbol{\hat{y}}$:
\begin{equation}\label{eq-importance}
  MACC(x_i, \boldsymbol{\hat{y}}) = \frac{1}{c}\sum^{c}_{j=1} abs(\rho ( x_i, \hat{y}_j)),
\end{equation}
where $\boldsymbol{\hat{y}} = \{\hat{y}_j\}^c_{j=1}$ denotes the model outputs and $\rho$ is the Pearson correlation coefficient~\cite{gregorutti2017correlation, Pearson}. 
{
Note that MACC measures the averaging linear correlation between $x_i$ and $\boldsymbol{\hat{y}}$. A larger MACC indicates that the change of $x_i$ can produce larger variations in $\boldsymbol{\hat{y}}$. }
We use the absolute values of correlation coefficients because we focus on the intensities instead of the directions of the correlations.

\begin{figure*}[t]
\centering
\begin{small}
\begin{tabular}{c}
\vspace{-2mm}
\subfigure[Mean Absolute Correlation Coefficients]{\includegraphics[width=0.85\textwidth]{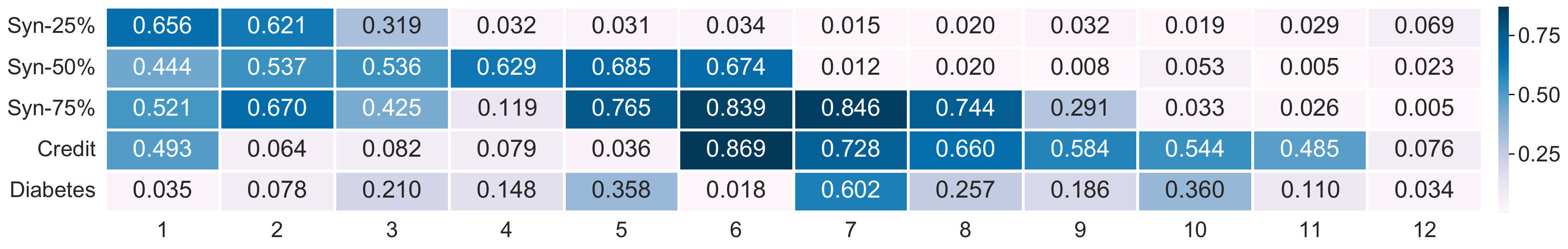}\label{subfig-per-feature-macc}}
 \\
\subfigure[$\ell_1$ loss]{\includegraphics[width=0.85\textwidth]{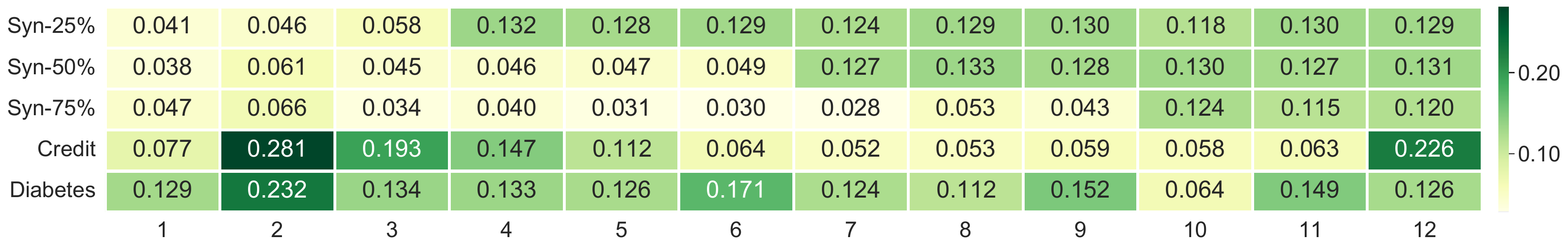}\label{subfig-per-feature-loss}}
\end{tabular}
\vspace{-5mm}
\caption{The (a) MACCs and  (b) $\ell_1$ losses of attack 1 \textit{w.r.t.} different features in different datasets. The horizontal axes denote all features in the synthesis datasets and the first 12 features in Credit and Diabetes.}
\label{fig-per-feature}
\end{small}
\vspace{-3mm}
\end{figure*}

\begin{figure}[t]
\centering
\begin{small}
\begin{tabular}{cc}
\multicolumn{2}{c}{\hspace{0mm} \includegraphics[height=3.5mm]{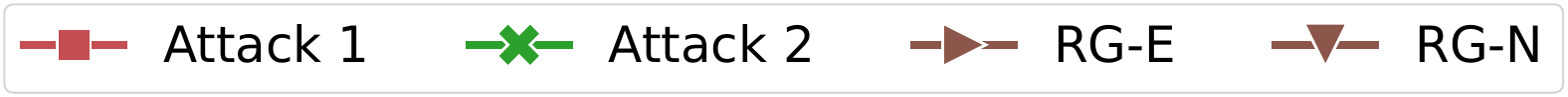}}
\vspace{-2mm}  \\
\hspace{-3mm}
\subfigure[Credit]{\includegraphics[width=0.5\columnwidth]{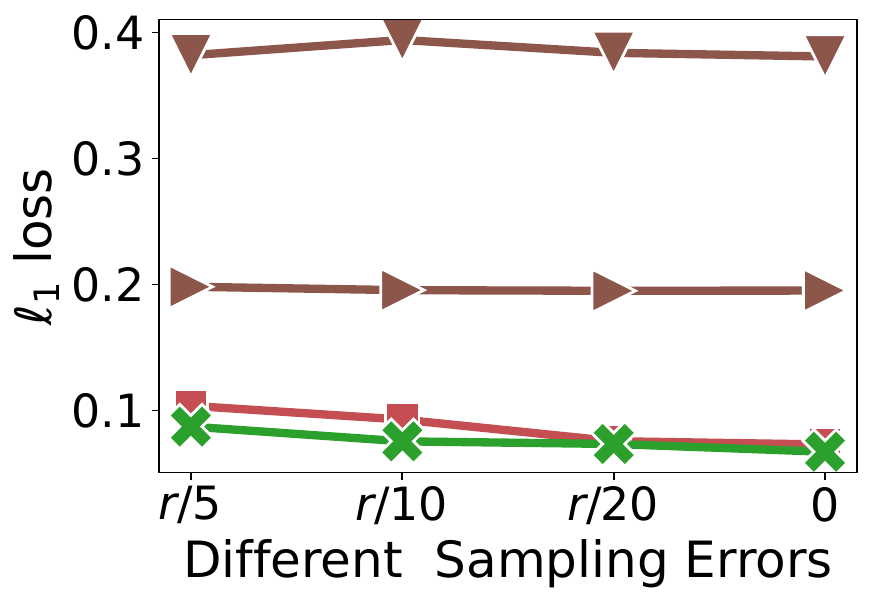}\label{subfig-sample-credit}}
&
\hspace{-3mm}
\subfigure[Diabetes]{\includegraphics[width=0.5\columnwidth]{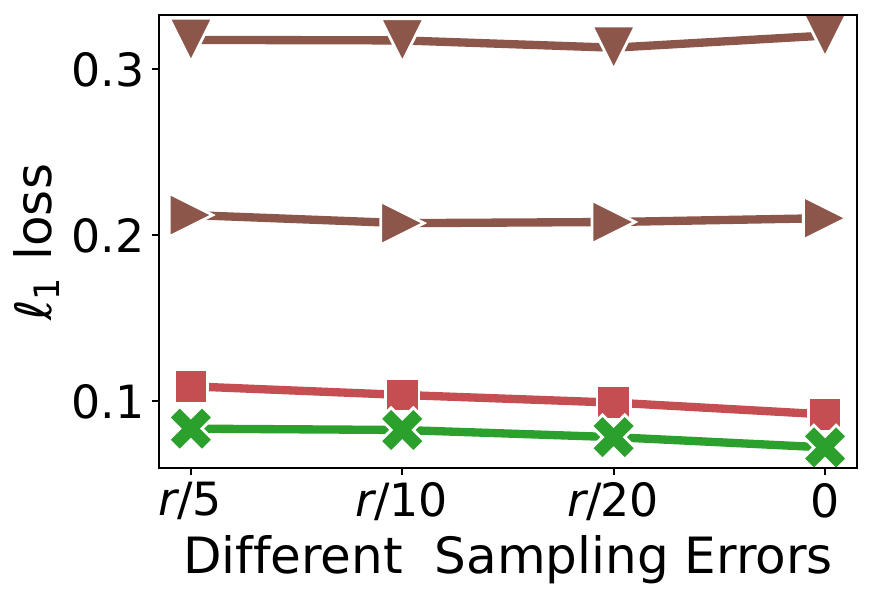}\label{subfig-sample-diabetes}}
\end{tabular}
\vspace{-5mm}
\caption{The performance of attack 1 $\&$ 2 \textit{w.r.t.} different sampling errors.}
\label{fig-sampling-error}
\end{small}
\vspace{-5mm}
\end{figure}

We first train an NN model on Google Cloud, then compute the MACCs of different features in $\mathcal{X}_{val}$ and show them in Fig.~\ref{subfig-per-feature-macc}. We then accordingly perform the proposed attacks on these models and show the $\ell_1$ loss per feature of adversary 1 in Fig.~\ref{subfig-per-feature-loss}. The losses and success rates per feature of adversary 2 follow a similar pattern and thus are omitted in this paper.
From Fig.~\ref{fig-per-feature}, we can see that the more important features (deeper colors in Fig.~\ref{subfig-per-feature-macc}) will be reconstructed with lower errors (lighter colors in Fig.~\ref{subfig-per-feature-loss}). 
This observation is consistent with the motivations of model interpretability methods: if a feature is more important, the variations of its value can cause a greater impact on the model outputs, thus producing a greater value on the model explanations.
But accordingly, its Shapley value will contain more information of this feature and can be accurately reconstructed in a larger probability. 
%


\subsection{Attack Performance \textit{w.r.t.} Different Shapley Sampling Errors}
As discussed in Section~\ref{subsec-attack-1} and \ref{subsec-attack-2}, the proposed attacks can be applied to the explanation methods with unknown sampling errors. To evaluate the robustness of the proposed algorithms, we first vary the Shapley sampling errors $\epsilon_s$ in $\{r/5, r/10, r/20, 0\}$ and accordingly generate Shapley values with different errors via the Vanilla method, then test the attack performance on these explanations. The $\delta$ in Eq.~\ref{eq-shap-sampling} is set to 0.1.
The test model is GBDT. 
We empirically set $\xi$ in Algorithm~\ref{alg-attack-2} to $\frac{r}{5}$ in all experiments, which corresponds to the real-world scenario that the adversary has no information about the sampling error. The numbers of queries in both attacks are set to 1000. The attack losses of adversary 1 and 2 on Credit are shown in Fig.~\ref{subfig-sample-credit}, with the success rates  $\{0.4685, 	0.4689, 	0.4758, 	0.4824 \}$ of adversary 2. The losses on Diabetes are shown in Fig.~\ref{subfig-sample-diabetes}, with the success rates  $\{0.4730, 	0.4921, 	0.4955, 	0.5057\}$ of adversary 2.

We make two observations from the results. First, the reconstruction accuracies of the proposed attacks slightly improve as the sampling errors decrease. This is reasonable because with the decrease of random noises, the information of the private features contained in Shapley values will increase, which is beneficial to the reconstruction algorithms. But the improvement is slight, demonstrating the robustness and efficiency of the proposed attacks even under large sampling errors. For justification, only 37 random permutations of features are needed to achieve the sampling error of $r/5$, whereas $2^n$ permutations are needed to achieve zero sampling errors.
Second, the losses of adversary 1 are slightly larger than the losses of adversary 2. The reason is that adversary 2 can only reconstruct roughly $50\%$ important features, while adversary 1 can reconstruct all features, among which the less important features contain more noise and thus reduce the overall attack accuracies of adversary 1. Nevertheless, both attacks can achieve far better accuracies than the random guess baselines.

\begin{figure*}[t]
\centering
\begin{small}
\begin{tabular}{cccc}
\multicolumn{2}{c}{\hspace{0mm} \includegraphics[height=3.5mm]{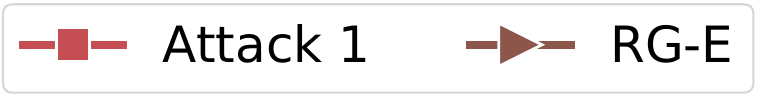}}
&
\multicolumn{2}{c}{\hspace{0mm} \includegraphics[height=3.5mm]{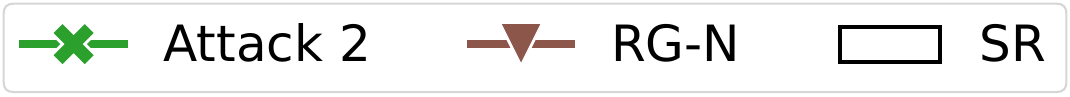}}
\vspace{-2mm}  \\
\hspace{-3mm}
\subfigure[Attack 1 on Adult]{\includegraphics[width=0.23\textwidth]{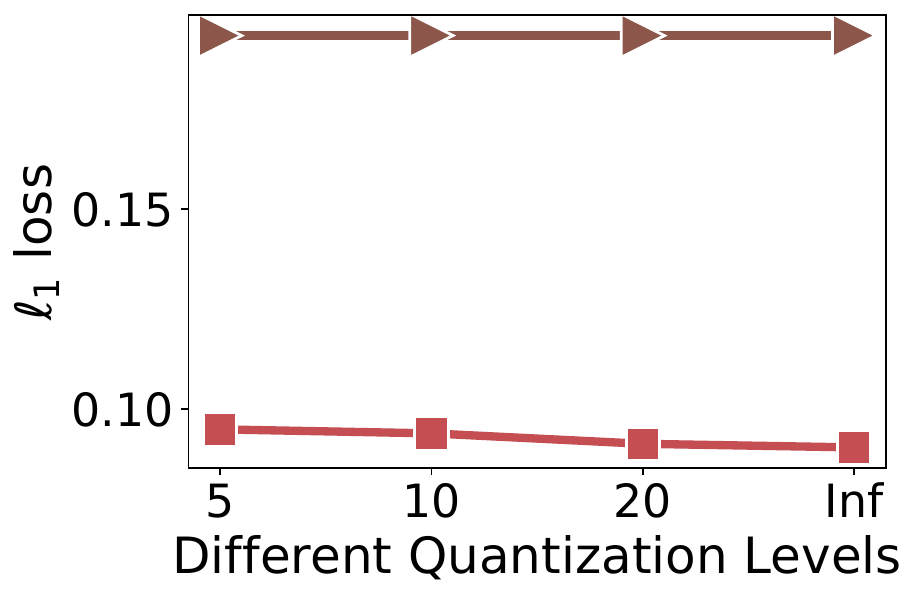}\label{subfig-quantization-start}}
&
\hspace{-3mm}
\subfigure[Attack 1 on Bank]{\includegraphics[width=0.225\textwidth]{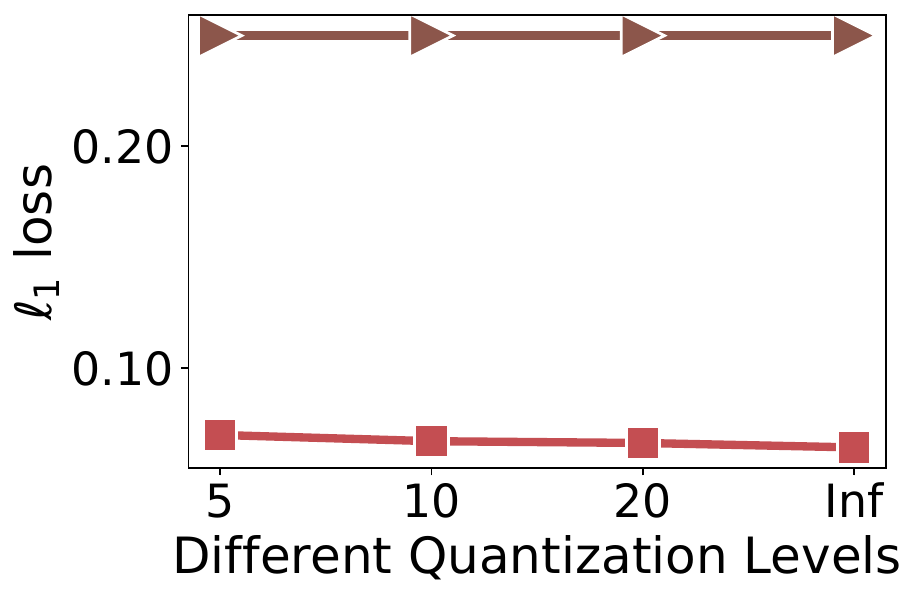}}
&
\hspace{-3mm}
\subfigure[Attack 2 on Adult]{\includegraphics[width=0.26\textwidth]{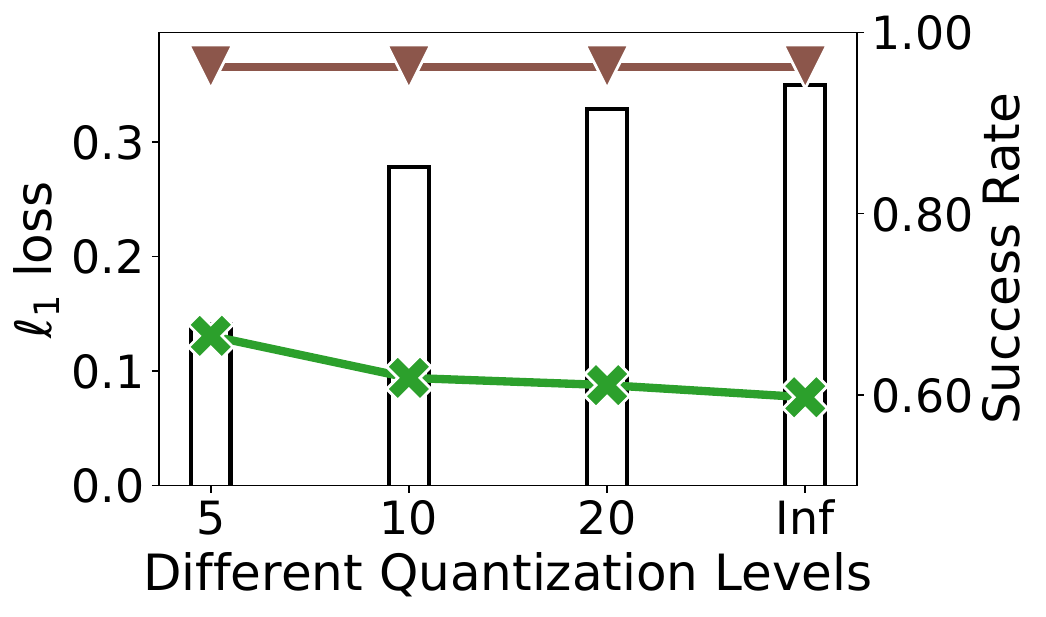}}
&
\hspace{-3mm}
\subfigure[Attack 2 on Bank]{\includegraphics[width=0.26\textwidth]{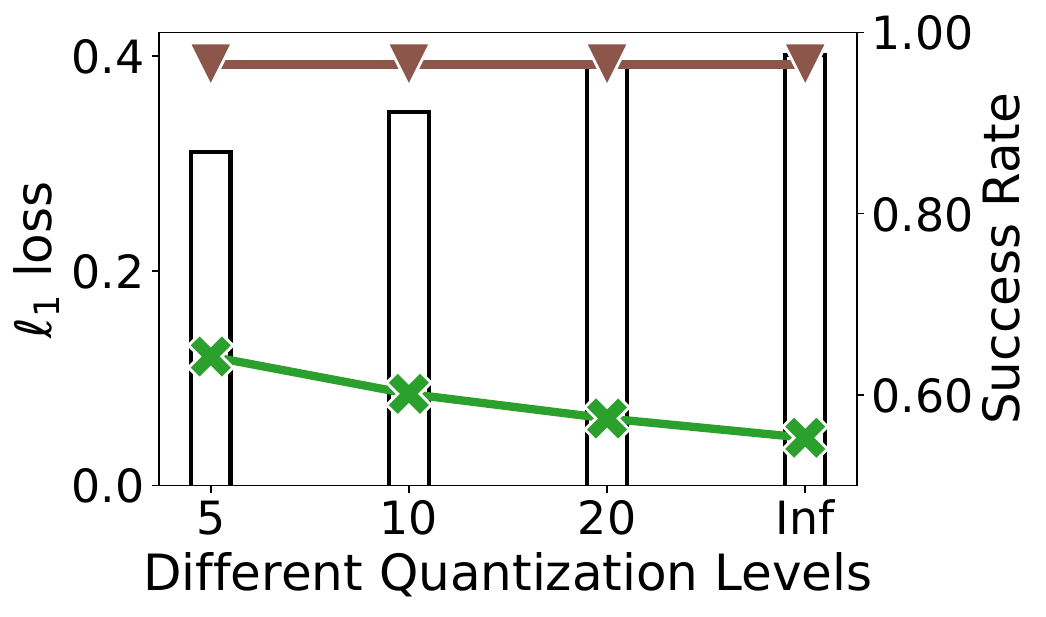}\label{subfig-quantization-end}}
\vspace{-2mm}  \\
\hspace{-3mm}
\subfigure[Attack 1 on Adult]{\includegraphics[width=0.23\textwidth]{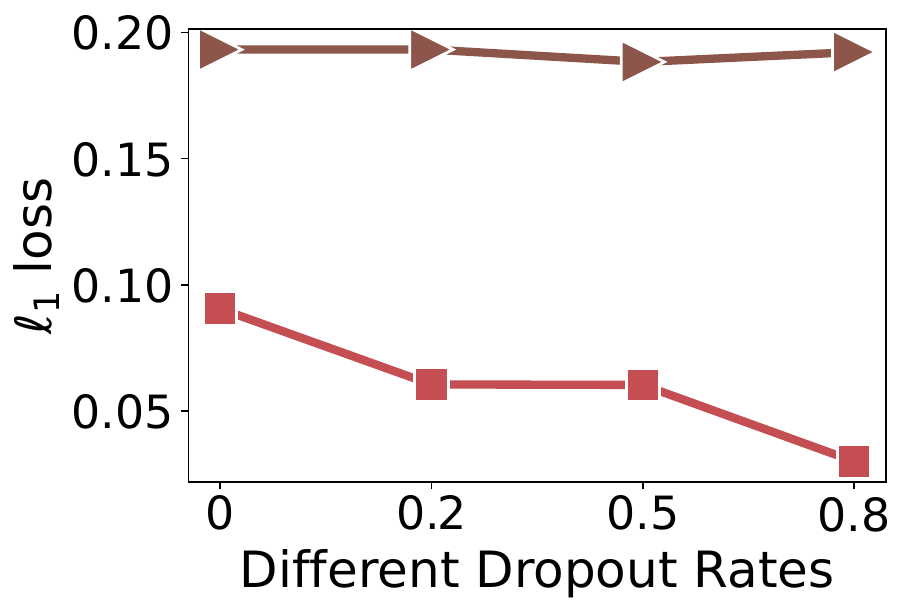}\label{subfig-dropout-start}}
&
\hspace{-3mm}
\subfigure[Attack 1 on Bank]{\includegraphics[width=0.225\textwidth]{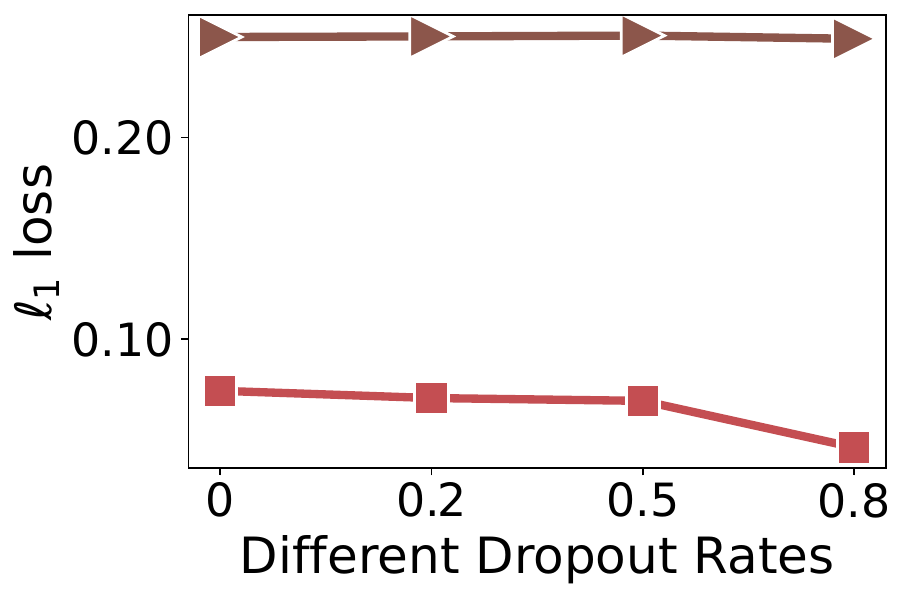}}
&
\hspace{-3mm}
\subfigure[Attack 2 on Adult]{\includegraphics[width=0.27\textwidth]{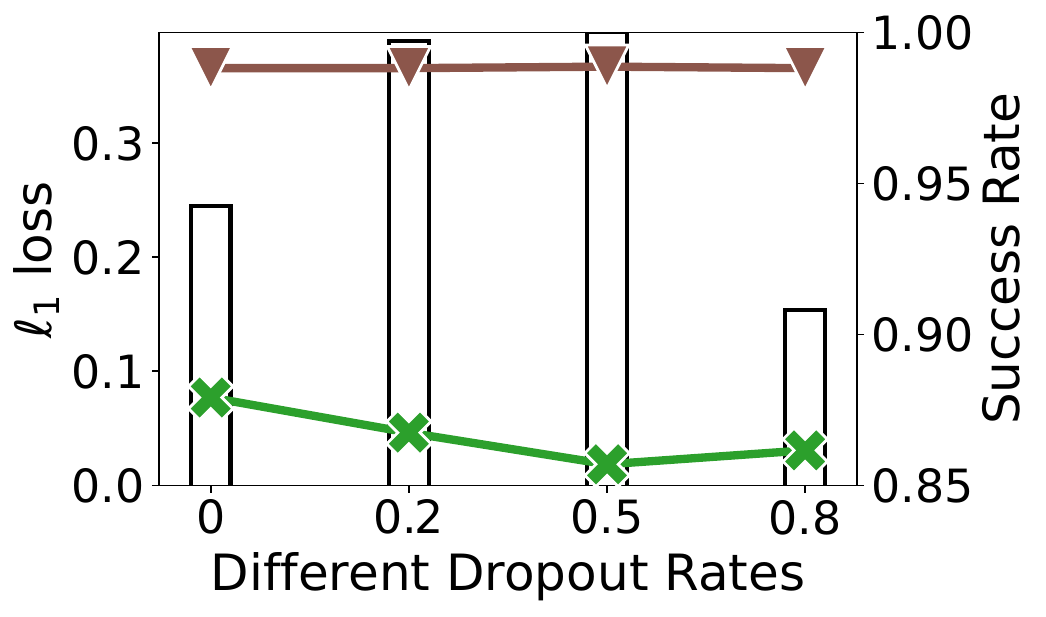}\label{subfig-drop-adult-2}}
&
\hspace{-3mm}
\subfigure[Attack 2 on Bank]{\includegraphics[width=0.27\textwidth]{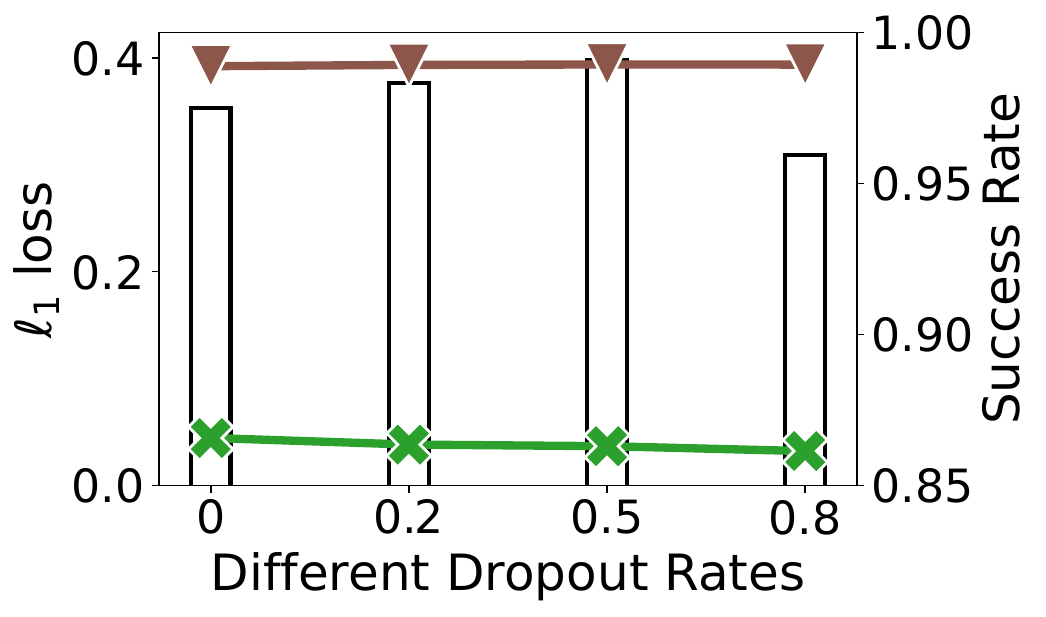}\label{subfig-drop-bank-2}\label{subfig-dropout-end}}
\end{tabular}
\vspace{-5mm}
\caption{The performance of attack 1 $\&$ 2 tested on (a)-(d) different quantization levels and (e)-(h) different dropout rates. The test model and platform are NN and Google Cloud.  The number of queries is set to 1600.}
\label{fig-defense}
\end{small}
\vspace{-3.5mm}
\end{figure*}

\section{Countermeasures}\label{sec-countermeasure}
{
\textbf{Reduce the Mutual Information between Inputs and Shapley Values.}
As discussed in Section~\ref{subsec-attack-1}, minimizing the mutual information $I(x_i; s_i)$ can help protect the privacy of $\boldsymbol{x}$, which has been used as a defense mechanism in previous works~\cite{wang2021privacy}.
Now we revisit the mutual information between $x_i$ and $s_i$
}:
\begin{equation}
  I(x_i; s_i)= H(s_i) - H(s_i|x_i)
 \stackrel{\text{(a)}}{\thickapprox}  H(s_i),
\end{equation}
where $H(s_i)$ denotes the entropy of $s_i$, and (a) follows because $s_i$ is a function of $x_i$ (see Eq.~\ref{eq-shap}). 
A way to reduce $I(x_i; s_i)$, i.e, $H(s_i)$, is quantizing the Shapley values. Specifically, we can first force the Shapley values of each feature to be one of $5/10/20$ different values, respectively, then evaluate the impact of this quantization strategy on the accuracies of the proposed attacks. The results are shown in Fig.~\ref{subfig-quantization-start}-\ref{subfig-quantization-end}.
We observe that the quantization strategy has little impact on the performance of attack 1.
Let $k$ denote the number of values defined in quantization, the space size of the target Shapley vector $\boldsymbol{s}$ is $k^n$, which could be far larger than $|\mathcal{S}_{aux}|$, indicating that the one-to-one correspondence between $\mathcal{X}_{aux}$ and $\mathcal{S}_{aux}$ still holds in a great probability.
Meanwhile, the quantization strategy can reduce the accuracies and success rates of attack 2, which is expected because the range of the candidate estimations for a feature is enlarged and thus leading to larger estimation errors (Eq.~\ref{eq-error-bound}).
Note that quantization can harm the utility of Shapley values, because two different input samples may produce the same explanation. Developing the privacy-preserving interpretability methods that  achieve a good utility-privacy trade-off could be a promising direction in future studies.

\vspace{0.5mm}\noindent
\textbf{Regularization.} Previous studies~\cite{luo2021feature, salem2018ml, melis2019exploiting, carlini2019secret, salem2018ml, tramer2016stealing} demonstrate that regularization techniques, including dropout and ensemble learning, can prevent the models from memorizing private inputs and thus mitigate the risks of information leakage.
In Section~\ref{subsec-exp-sizes}, we discuss the attack performance on different models and conclude that ensemble models can reduce the attack accuracies to a limited degree. In this section, we discuss the impact of the dropout strategy~\cite{srivastava2014dropout} in neural networks on the accuracies of the proposed attacks. In the experiments, we first add a dropout layer after each hidden layer and vary the dropout rate in $\{0.2, 0.5, 0.8\}$, then evaluate the corresponding attack performance.
From Fig.~\ref{subfig-dropout-start}-\ref{subfig-dropout-end}, we make an interesting observation that instead of harming the attack accuracies, the Dropout strategy improves the attack accuracies, which is entirely different from previous studies~\cite{luo2021feature, salem2018ml, melis2019exploiting, carlini2019secret}.
The reason is that Dropout prevents NN models from overfitting by smoothing the decision boundaries during training, which provides more advantages to the proposed attacks.
Note that from Fig.~\ref{subfig-drop-adult-2} and \ref{subfig-drop-bank-2}, we see the success rates of attack 2 drop when the dropout rate is 0.8. The reason is that under this dropout rate, the NN models are underfitting with decision boundaries containing randomness produced in the parameter initialization, thus reducing the linearity between inputs and outputs.
However, this randomness causes little impact on attack 1, because the MLP attack model used in attack 1 can learn different decision boundaries with a desired accuracy~\cite{hornik1989multilayer}.


\vspace{0.5mm}\noindent
{\textbf{Reduce the Dimension of Shapley Values.} Because the number of Shapley values \textit{w.r.t.} a class is equal to the number of input features (see Fig.~\ref{fig-attack-illustration}), one intuitive defense is to reduce the number of Shapley values released. 
Note that the variance of Shapley values, instead of the value intensity, indicates the importance of features (see Eq.~\ref{eq-mutual-information-duction-4}), thus simply releasing the largest $k$ Shapley values may harm the utility of explanations because the important features in some samples may have small Shapley values. 
Instead, we can empirically choose the top $k$ important features based on their Shapley variances and only release the Shapley values of these $k$ features as a defense.
It is worth noting that this defense method causes little impact on the accuracy of Attack 2 because different features are reconstructed independently based on the corresponding Shapley values (Algorithm~\ref{alg-attack-2}). Therefore, we mainly test the performance of Attack 1 under this defense. In the experiment, we only release $k$ Shapley values and recover the $k$ corresponding features. The experimental setting is the same as in Fig.~\ref{fig-defense}. 
When the numbers of released Shapley values vary in $\{20\%n, 50\%n, n\}$, the averaging $\ell_1$ losses of Adult and Bank are $\{0.0824, 0.0909, 0.0905\}$ and $\{0.0546, 0.0619, 0.0643\}$, respectively. The results are consistent with our analysis: the Shapley values of more important features leak more private information (Section~\ref{subsec-a1-motivation}) since the models tend to build linear connections between these features and their predictions (Section~\ref{subsubsec-attack2}).
In summary, releasing only $k$ Shapley values can reduce the success rates of the proposed attacks to at most $k/n$ but impact little on the accuracy of the recovered features. }

\vspace{0.5mm}\noindent
\textbf{Differential Privacy (DP).} DP~\cite{xiao2010differential} is a state-of-the-art privacy-preserving mechanism that can protect the private inputs via a rigorous theoretical guarantee. By requiring that the change of an arbitrary sample in the input dataset should not be reflected in the outputs, DP can prevent the adversaries from inferring the information of private inputs via observing the outputs.
However, DP does not apply to local interpretability methods and thus can not be employed as a defense against the proposed attacks. Specifically, for two arbitrary private samples $\boldsymbol{x}^i$ and $\boldsymbol{x}^j$ with $\boldsymbol{x}^i\neq \boldsymbol{x}^j$, local interpretability methods compute two explanations $\boldsymbol{s}^i$ and $\boldsymbol{s}^j$.
To satisfy the requirements of DP, $\boldsymbol{s}^i$ and $\boldsymbol{s}^j$ should be randomized to be indistinguishable, which harms the utility of Shapley values since different samples will produce nearly the same explanations. Therefore, DP is inapplicable to the current problem.

\section{Related Work}\label{sec-related-work}

\textbf{Model Interpretability.} 
%
The lack of theoretical techniques for tracking data flows in complicated black-box models  fosters plenty of studies on model interpretability, which aims to explain how these models produce their predictions~\cite{rudin2019stop}. 
The model interpretability methods can be classified into two categories: \textit{global interpretability} for computing the global importance of different features~\cite{fisher2019all, covert2020SAGEGlobalShapely, datta2016algorithmic}, 
and \textit{local interpretability} for estimating the importance of features in a target sample~\cite{ribeiro2016LIME, lundberg2017SHAP, shrikumar2017DeepLIFT}. 
%
In this paper, we focus on the local methods, which can be further classified into Shapley value-based methods and heuristic methods. The \textit{heuristic methods} aim to linearize the model decision boundaries near the target samples and use the linear weights of different features as their importance values, e.g., LIME~\cite{ribeiro2016LIME} and gradient-based methods~\cite{shrikumar2017DeepLIFT, ancona2017towards, selvaraju2017grad}.
Although the heuristic methods are computationally efficient, few theoretical analyses are provided to justify their robustness and error bounds. On the other hand, \textit{Shapley value-based methods}~\cite{lundberg2017SHAP, covert2020SAGEGlobalShapely, datta2016algorithmic, chen2018shapley, maleki2013bounding, vstrumbelj2014explaining, chen2018learning} can theoretically provide the desirable properties for ML explanations and are thus adopted by the leading MLaaS platforms for explaining tabular data~\cite{Google, Microsoft, IBM}.
%

\vspace{0.5mm}\noindent
\textbf{Attack Algorithms on Machine Learning.}
Many studies have revealed the privacy risks suffered by machine learning models, such as membership inference~\cite{salem2018ml, shokri2017membership, nasr2019comprehensive, shokri2021privacy}, property inference~\cite{ganju2018property, melis2019exploiting, ateniese2015hacking}, model extraction~\cite{aivodji2020model, tramer2016stealing, milli2019model}  and feature inference~\cite{luo2021feature, wu2016methodology, fredrikson2015model, zhu2019deep}. \textit{Membership inference} aims to determine whether a specific sample is in the training dataset or not, and \textit{property inference} aims to restore the statistics of training datasets. Because these two types of attacks are supervised tasks, an auxiliary dataset is necessary for fulfilling the training of attack models. \textit{Model extraction} aims to reconstruct the parameters of black-box models from their predictions.
Nevertheless, few studies have considered the privacy risks of model interpretability methods.
%
Although some newest studies focus on membership inference~\cite{shokri2021privacy} or neural network extraction~\cite{aivodji2020model, milli2019model} based on the local explanations produced by heuristic interpretability methods, their settings and tasks are entirely different from ours.
In this paper, we focus on \textit{feature inference} attacks on Shapley values, which are more practical and impactive to the real-world applications. To our best knowledge, this is also the first paper that considers the privacy risks in Shapley values.

\vspace{0.5mm}\noindent
\textbf{Defense Mechanisms against Current Attacks.}
Since the efficiency of current attacks relies on the memorization nature of ML models~\cite{carlini2019secret, luo2021feature}, most defense mechanisms focus on preventing models from memorizing sensitive data via noise injections.
For example, using a lower granularity of model outputs can reduce the information of private features contained in them and accordingly decrease the accuracies of membership inference~\cite{shokri2017membership} and model extraction~\cite{tramer2016stealing}.
The regularization methods, such as dropout in NN~\cite{luo2021feature, salem2018ml, melis2019exploiting, carlini2019secret} and ensemble learning~\cite{salem2018ml, tramer2016stealing}, can reduce the information of training datasets memorized by the model. 
%
Differential privacy~\cite{xiao2010differential} is one of the most popular defense mechanisms because it provides a rigorous theoretical guarantee of protecting the private model inputs~\cite{luo2021feature, carlini2019secret, tramer2016stealing, melis2019exploiting}.
However, these defenses can not apply to the proposed attacks, because we focus on reconstructing the private inputs in the prediction phase which have never been seen and thus memorized by the model. 
%
%
{\cite{wang2021privacy} proposes a defense mechanism by  reducing the mutual information between the plaintext outputs and private data, which is similar in spirit to our analysis. However, the method in \cite{wang2021privacy} can not be directly used in our setting. First, \cite{wang2021privacy} aims to change the learning objectives via mutual information during model training, whereas the model is trained and fixed in the computation of Shapley values.
Second, the mutual information for privacy preserving in \cite{wang2021privacy} is loosely related to the primary task and thus can be minimized, whereas in our case, minimizing the mutual information between Shapley values and features can harm the utility of explanations. Nevertheless, \cite{wang2021privacy} still provides some insights on mitigating the proposed attacks.}

\section{Discussion}\label{sec-discussion}
{
\textbf{The Choice of Datasets.} As shown in Fig.~\ref{fig-loss-sizes} and Tab.~\ref{tb-success-sizes-attack2}, different choices of experimental datasets impact little on the attack performance. First, the number of samples in the training datasets has no relation to the proposed attacks since we focus on reconstructing private features in the prediction phase.
Then, the number of classes also has limited impact on the proposed attacks because we use only one explanation vector \textit{w.r.t.} the class with the highest confidence score (see Section~\ref{sec-problem}).
In addition, using larger feature sets may reduce the success rates of Attack 2, which, however, should be attributed to the model behaviors, because Attack 2 can only reconstruct those features that have relatively large influence on the model outputs (see Section~\ref{subsubsec-attack2}).}

\vspace{0.5mm}\noindent
{
\textbf{The Influencing Factors of the Attacks.}
Our experiments reveal two main factors that impact the attack performance. 
First, from the results of Section~\ref{subsec-exp-sizes}, we conclude that in general, NN $>$ SVM $>$ RF $>$ GBDT \textit{w.r.t.} the vulnerability to attacks. NN is most vulnerable because the non-linear activations, e.g., ReLU and Sigmoid, can behave in very linear ways for the important features~\cite{goodfellow2014explaining}, leading to a linear correlation between features and their Shapley values.
Second, the results in Section~\ref{subsec-exp-feature} show that the important features, which have close relations with model outputs, are more vulnerable to the proposed attacks. Eq.~\ref{eq-mutual-information-duction-4} also shows that the variances of Shapley values corresponding to important features are larger than that of less important features.
These results provide some insights that can be utilized by future studies to design defenses against privacy attacks on Shapley values.
}

\section{Conclusion}\label{sec-conclusion}
In this paper, we focus on the privacy risks suffered in the Shapley value-based model explanation methods.
We first analyze the connections between private model inputs and their Shapley values in an information-theoretical perspective, then accordingly propose two feature inference attacks under different settings. 
Extensive experiments performed on three leading MLaaS platforms demonstrate the effectiveness and robustness of the proposed attacks, highlighting the necessity of
developing privacy-preserving model interpretability methods in future studies.

\begin{acks}
We would like to thank Yuncheng Wu, Xiangqi Zhu, and
Xiaochen Zhu for helpful discussions and feedback.
This work was supported by the Ministry of Education, Singapore (Number MOE2018-T2-2-091) and A*STAR, Singapore (Number A19E3b0099). Any opinions, findings and conclusions or recommendations expressed in this material are those of the authors and do not reflect the views of the funding agencies.

\end{acks}



\bibliographystyle{ACM-Reference-Format.bst}
\balance
\bibliography{references.bib}


\begin{thebibliography}{70}


\ifx \showCODEN    \undefined \def \showCODEN     #1{\unskip}     \fi
\ifx \showDOI      \undefined \def \showDOI       #1{#1}\fi
\ifx \showISBNx    \undefined \def \showISBNx     #1{\unskip}     \fi
\ifx \showISBNxiii \undefined \def \showISBNxiii  #1{\unskip}     \fi
\ifx \showISSN     \undefined \def \showISSN      #1{\unskip}     \fi
\ifx \showLCCN     \undefined \def \showLCCN      #1{\unskip}     \fi
\ifx \shownote     \undefined \def \shownote      #1{#1}          \fi
\ifx \showarticletitle \undefined \def \showarticletitle #1{#1}   \fi
\ifx \showURL      \undefined \def \showURL       {\relax}        \fi
\providecommand\bibfield[2]{#2}
\providecommand\bibinfo[2]{#2}
\providecommand\natexlab[1]{#1}
\providecommand\showeprint[2][]{arXiv:#2}

\bibitem[Agarwal et~al\mbox{.}(2021)]%
        {agarwal2021neural}
\bibfield{author}{\bibinfo{person}{Rishabh Agarwal}, \bibinfo{person}{Levi
  Melnick}, \bibinfo{person}{Nicholas Frosst}, \bibinfo{person}{Xuezhou Zhang},
  \bibinfo{person}{Benjamin~J. Lengerich}, \bibinfo{person}{Rich Caruana},
  {and} \bibinfo{person}{Geoffrey~E. Hinton}.} \bibinfo{year}{2021}\natexlab{}.
\newblock \showarticletitle{Neural Additive Models: Interpretable Machine
  Learning with Neural Nets}. In \bibinfo{booktitle}{\emph{Advances in Neural
  Information Processing Systems 34: Annual Conference on Neural Information
  Processing Systems 2021, NeurIPS 2021, December 6-14, 2021, virtual}}.
  \bibinfo{pages}{4699--4711}.
\newblock


\bibitem[AI(2022)]%
        {IBM}
\bibfield{author}{\bibinfo{person}{IBM Research~Trusted AI}.}
  \bibinfo{year}{2022}\natexlab{}.
\newblock \bibinfo{title}{AI Explainability 360 - Resources}.
\newblock \bibinfo{howpublished}{\url{http://aix360.mybluemix.net/resources}}.
\newblock
\newblock
\shownote{Online; accessed 24-March-2022}.


\bibitem[A{\"{\i}}vodji et~al\mbox{.}(2020)]%
        {aivodji2020model}
\bibfield{author}{\bibinfo{person}{Ulrich A{\"{\i}}vodji},
  \bibinfo{person}{Alexandre Bolot}, {and} \bibinfo{person}{S{\'{e}}bastien
  Gambs}.} \bibinfo{year}{2020}\natexlab{}.
\newblock \showarticletitle{Model extraction from counterfactual explanations}.
\newblock \bibinfo{journal}{\emph{CoRR}}  \bibinfo{volume}{abs/2009.01884}
  (\bibinfo{year}{2020}).
\newblock
\showeprint[arXiv]{2009.01884}
\urldef\tempurl%
\url{https://arxiv.org/abs/2009.01884}
\showURL{%
\tempurl}


\bibitem[Ancona et~al\mbox{.}(2018)]%
        {ancona2017towards}
\bibfield{author}{\bibinfo{person}{Marco Ancona}, \bibinfo{person}{Enea
  Ceolini}, \bibinfo{person}{Cengiz {\"{O}}ztireli}, {and}
  \bibinfo{person}{Markus Gross}.} \bibinfo{year}{2018}\natexlab{}.
\newblock \showarticletitle{Towards better understanding of gradient-based
  attribution methods for Deep Neural Networks}. In
  \bibinfo{booktitle}{\emph{6th International Conference on Learning
  Representations, {ICLR} 2018, Vancouver, BC, Canada, April 30 - May 3, 2018,
  Conference Track Proceedings}}. \bibinfo{publisher}{OpenReview.net}.
\newblock


\bibitem[Ancona et~al\mbox{.}(2019)]%
        {ancona2019explaining}
\bibfield{author}{\bibinfo{person}{Marco Ancona}, \bibinfo{person}{Cengiz
  {\"{O}}ztireli}, {and} \bibinfo{person}{Markus~H. Gross}.}
  \bibinfo{year}{2019}\natexlab{}.
\newblock \showarticletitle{Explaining Deep Neural Networks with a Polynomial
  Time Algorithm for Shapley Value Approximation}. In
  \bibinfo{booktitle}{\emph{Proceedings of the 36th International Conference on
  Machine Learning, {ICML} 2019, 9-15 June 2019, Long Beach, California,
  {USA}}}, Vol.~\bibinfo{volume}{97}. \bibinfo{publisher}{{PMLR}},
  \bibinfo{pages}{272--281}.
\newblock


\bibitem[Arya et~al\mbox{.}(2019)]%
        {arya2019IBM}
\bibfield{author}{\bibinfo{person}{Vijay Arya}, \bibinfo{person}{Rachel K.~E.
  Bellamy}, \bibinfo{person}{Pin{-}Yu Chen}, \bibinfo{person}{Amit Dhurandhar},
  \bibinfo{person}{Michael Hind}, \bibinfo{person}{Samuel~C. Hoffman},
  \bibinfo{person}{Stephanie Houde}, \bibinfo{person}{Q.~Vera Liao},
  \bibinfo{person}{Ronny Luss}, \bibinfo{person}{Aleksandra Mojsilovic},
  \bibinfo{person}{Sami Mourad}, \bibinfo{person}{Pablo Pedemonte},
  \bibinfo{person}{Ramya Raghavendra}, \bibinfo{person}{John~T. Richards},
  \bibinfo{person}{Prasanna Sattigeri}, \bibinfo{person}{Karthikeyan
  Shanmugam}, \bibinfo{person}{Moninder Singh}, \bibinfo{person}{Kush~R.
  Varshney}, \bibinfo{person}{Dennis Wei}, {and} \bibinfo{person}{Yunfeng
  Zhang}.} \bibinfo{year}{2019}\natexlab{}.
\newblock \showarticletitle{One Explanation Does Not Fit All: {A} Toolkit and
  Taxonomy of {AI} Explainability Techniques}.
\newblock \bibinfo{journal}{\emph{CoRR}}  \bibinfo{volume}{abs/1909.03012}
  (\bibinfo{year}{2019}).
\newblock
\showeprint[arXiv]{1909.03012}
\urldef\tempurl%
\url{http://arxiv.org/abs/1909.03012}
\showURL{%
\tempurl}


\bibitem[Ateniese et~al\mbox{.}(2015)]%
        {ateniese2015hacking}
\bibfield{author}{\bibinfo{person}{Giuseppe Ateniese},
  \bibinfo{person}{Luigi~V. Mancini}, \bibinfo{person}{Angelo Spognardi},
  \bibinfo{person}{Antonio Villani}, \bibinfo{person}{Domenico Vitali}, {and}
  \bibinfo{person}{Giovanni Felici}.} \bibinfo{year}{2015}\natexlab{}.
\newblock \showarticletitle{Hacking smart machines with smarter ones: How to
  extract meaningful data from machine learning classifiers}.
\newblock \bibinfo{journal}{\emph{International Journal of Security and
  Networks}} \bibinfo{volume}{10}, \bibinfo{number}{3} (\bibinfo{year}{2015}),
  \bibinfo{pages}{137--150}.
\newblock


\bibitem[Azure(2022)]%
        {Microsoft}
\bibfield{author}{\bibinfo{person}{Microsoft Azure}.}
  \bibinfo{year}{2022}\natexlab{}.
\newblock \bibinfo{title}{Model interpretability in Azure Machine Learning}.
\newblock
  \bibinfo{howpublished}{\url{https://docs.microsoft.com/en-us/azure/machine-learning/how-to-machine-learning-interpretability}}.
\newblock
\newblock
\shownote{Online; accessed 24-March-2022}.


\bibitem[Biau et~al\mbox{.}(2016)]%
        {biau2019neural}
\bibfield{author}{\bibinfo{person}{G{\'{e}}rard Biau}, \bibinfo{person}{Erwan
  Scornet}, {and} \bibinfo{person}{Johannes Welbl}.}
  \bibinfo{year}{2016}\natexlab{}.
\newblock \showarticletitle{Neural Random Forests}.
\newblock \bibinfo{journal}{\emph{CoRR}}  \bibinfo{volume}{abs/1604.07143}
  (\bibinfo{year}{2016}).
\newblock
\showeprint[arXiv]{1604.07143}
\urldef\tempurl%
\url{http://arxiv.org/abs/1604.07143}
\showURL{%
\tempurl}


\bibitem[Boser et~al\mbox{.}(1992)]%
        {boser1992training}
\bibfield{author}{\bibinfo{person}{Bernhard~E. Boser},
  \bibinfo{person}{Isabelle Guyon}, {and} \bibinfo{person}{Vladimir Vapnik}.}
  \bibinfo{year}{1992}\natexlab{}.
\newblock \showarticletitle{A Training Algorithm for Optimal Margin
  Classifiers}. In \bibinfo{booktitle}{\emph{Proceedings of the Fifth Annual
  {ACM} Conference on Computational Learning Theory, {COLT} 1992, Pittsburgh,
  PA, USA, July 27-29, 1992}}. \bibinfo{publisher}{{ACM}},
  \bibinfo{pages}{144--152}.
\newblock


\bibitem[Breiman(2001)]%
        {breiman2001random}
\bibfield{author}{\bibinfo{person}{Leo Breiman}.}
  \bibinfo{year}{2001}\natexlab{}.
\newblock \showarticletitle{Random forests}.
\newblock \bibinfo{journal}{\emph{Machine learning}} \bibinfo{volume}{45},
  \bibinfo{number}{1} (\bibinfo{year}{2001}), \bibinfo{pages}{5--32}.
\newblock


\bibitem[Carlini et~al\mbox{.}(2019)]%
        {carlini2019secret}
\bibfield{author}{\bibinfo{person}{Nicholas Carlini}, \bibinfo{person}{Chang
  Liu}, \bibinfo{person}{{\'{U}}lfar Erlingsson}, \bibinfo{person}{Jernej Kos},
  {and} \bibinfo{person}{Dawn Song}.} \bibinfo{year}{2019}\natexlab{}.
\newblock \showarticletitle{The Secret Sharer: Evaluating and Testing
  Unintended Memorization in Neural Networks}. In
  \bibinfo{booktitle}{\emph{28th {USENIX} Security Symposium, {USENIX} Security
  2019, Santa Clara, CA, USA, August 14-16, 2019}}.
  \bibinfo{publisher}{{USENIX} Association}, \bibinfo{pages}{267--284}.
\newblock


\bibitem[Chen et~al\mbox{.}(2018)]%
        {chen2018learning}
\bibfield{author}{\bibinfo{person}{Jianbo Chen}, \bibinfo{person}{Le Song},
  \bibinfo{person}{Martin~J. Wainwright}, {and} \bibinfo{person}{Michael~I.
  Jordan}.} \bibinfo{year}{2018}\natexlab{}.
\newblock \showarticletitle{Learning to Explain: An Information-Theoretic
  Perspective on Model Interpretation}. In
  \bibinfo{booktitle}{\emph{Proceedings of the 35th International Conference on
  Machine Learning, {ICML} 2018, Stockholmsm{\"{a}}ssan, Stockholm, Sweden,
  July 10-15, 2018}}, Vol.~\bibinfo{volume}{80}. \bibinfo{publisher}{{PMLR}},
  \bibinfo{pages}{882--891}.
\newblock


\bibitem[Chen et~al\mbox{.}(2019)]%
        {chen2018shapley}
\bibfield{author}{\bibinfo{person}{Jianbo Chen}, \bibinfo{person}{Le Song},
  \bibinfo{person}{Martin~J. Wainwright}, {and} \bibinfo{person}{Michael~I.
  Jordan}.} \bibinfo{year}{2019}\natexlab{}.
\newblock \showarticletitle{L-Shapley and C-Shapley: Efficient Model
  Interpretation for Structured Data}. In \bibinfo{booktitle}{\emph{7th
  International Conference on Learning Representations, {ICLR} 2019, New
  Orleans, LA, USA, May 6-9, 2019}}. \bibinfo{publisher}{OpenReview.net}.
\newblock


\bibitem[Cloud(2022)]%
        {Google}
\bibfield{author}{\bibinfo{person}{Google Cloud}.}
  \bibinfo{year}{2022}\natexlab{}.
\newblock \bibinfo{title}{Introduction to AI Explanations for AI Platform}.
\newblock
  \bibinfo{howpublished}{\url{https://cloud.google.com/ai-platform/prediction/docs/ai-explanations/overview}}.
\newblock
\newblock
\shownote{Online; accessed 24-March-2022}.


\bibitem[Covert and Lee(2021)]%
        {covert2021improving}
\bibfield{author}{\bibinfo{person}{Ian Covert} {and} \bibinfo{person}{Su{-}In
  Lee}.} \bibinfo{year}{2021}\natexlab{}.
\newblock \showarticletitle{Improving KernelSHAP: Practical Shapley Value
  Estimation Using Linear Regression}. In \bibinfo{booktitle}{\emph{The 24th
  International Conference on Artificial Intelligence and Statistics, {AISTATS}
  2021, April 13-15, 2021, Virtual Event}}, Vol.~\bibinfo{volume}{130}.
  \bibinfo{publisher}{{PMLR}}, \bibinfo{pages}{3457--3465}.
\newblock


\bibitem[Covert et~al\mbox{.}(2020)]%
        {covert2020SAGEGlobalShapely}
\bibfield{author}{\bibinfo{person}{Ian Covert}, \bibinfo{person}{Scott~M.
  Lundberg}, {and} \bibinfo{person}{Su{-}In Lee}.}
  \bibinfo{year}{2020}\natexlab{}.
\newblock \showarticletitle{Understanding Global Feature Contributions With
  Additive Importance Measures}. In \bibinfo{booktitle}{\emph{Advances in
  Neural Information Processing Systems 33: Annual Conference on Neural
  Information Processing Systems 2020, NeurIPS 2020, December 6-12, 2020,
  virtual}}. \bibinfo{pages}{17212--17223}.
\newblock


\bibitem[Datta et~al\mbox{.}(2016)]%
        {datta2016algorithmic}
\bibfield{author}{\bibinfo{person}{Anupam Datta}, \bibinfo{person}{Shayak Sen},
  {and} \bibinfo{person}{Yair Zick}.} \bibinfo{year}{2016}\natexlab{}.
\newblock \showarticletitle{Algorithmic Transparency via Quantitative Input
  Influence: Theory and Experiments with Learning Systems}. In
  \bibinfo{booktitle}{\emph{{IEEE} Symposium on Security and Privacy, {SP}
  2016, San Jose, CA, USA, May 22-26, 2016}}. \bibinfo{publisher}{{IEEE}
  Computer Society}, \bibinfo{pages}{598--617}.
\newblock


\bibitem[Dua and Graff(2017)]%
        {UCI}
\bibfield{author}{\bibinfo{person}{Dheeru Dua} {and} \bibinfo{person}{Casey
  Graff}.} \bibinfo{year}{2017}\natexlab{}.
\newblock \bibinfo{title}{{UCI} Machine Learning Repository}.
\newblock
\newblock
\urldef\tempurl%
\url{http://archive.ics.uci.edu/ml}
\showURL{%
\tempurl}


\bibitem[Fisher et~al\mbox{.}(2019)]%
        {fisher2019all}
\bibfield{author}{\bibinfo{person}{Aaron Fisher}, \bibinfo{person}{Cynthia
  Rudin}, {and} \bibinfo{person}{Francesca Dominici}.}
  \bibinfo{year}{2019}\natexlab{}.
\newblock \showarticletitle{All Models are Wrong, but Many are Useful: Learning
  a Variable's Importance by Studying an Entire Class of Prediction Models
  Simultaneously}.
\newblock \bibinfo{journal}{\emph{J. Mach. Learn. Res.}}  \bibinfo{volume}{20}
  (\bibinfo{year}{2019}), \bibinfo{pages}{177:1--177:81}.
\newblock


\bibitem[Fredrikson et~al\mbox{.}(2015)]%
        {fredrikson2015model}
\bibfield{author}{\bibinfo{person}{Matt Fredrikson}, \bibinfo{person}{Somesh
  Jha}, {and} \bibinfo{person}{Thomas Ristenpart}.}
  \bibinfo{year}{2015}\natexlab{}.
\newblock \showarticletitle{Model Inversion Attacks that Exploit Confidence
  Information and Basic Countermeasures}. In
  \bibinfo{booktitle}{\emph{Proceedings of the 22nd {ACM} {SIGSAC} Conference
  on Computer and Communications Security, Denver, CO, USA, October 12-16,
  2015}}. \bibinfo{publisher}{{ACM}}, \bibinfo{pages}{1322--1333}.
\newblock


\bibitem[Friedman(2001)]%
        {friedman2001greedy}
\bibfield{author}{\bibinfo{person}{Jerome~H Friedman}.}
  \bibinfo{year}{2001}\natexlab{}.
\newblock \showarticletitle{Greedy function approximation: a gradient boosting
  machine}.
\newblock \bibinfo{journal}{\emph{Annals of statistics}}
  (\bibinfo{year}{2001}), \bibinfo{pages}{1189--1232}.
\newblock


\bibitem[Fujimoto et~al\mbox{.}(2006)]%
        {fujimoto2006axiomatic}
\bibfield{author}{\bibinfo{person}{Katsushige Fujimoto}, \bibinfo{person}{Ivan
  Kojadinovic}, {and} \bibinfo{person}{Jean-Luc Marichal}.}
  \bibinfo{year}{2006}\natexlab{}.
\newblock \showarticletitle{Axiomatic characterizations of probabilistic and
  cardinal-probabilistic interaction indices}.
\newblock \bibinfo{journal}{\emph{Games and Economic Behavior}}
  \bibinfo{volume}{55}, \bibinfo{number}{1} (\bibinfo{year}{2006}),
  \bibinfo{pages}{72--99}.
\newblock


\bibitem[Ganju et~al\mbox{.}(2018)]%
        {ganju2018property}
\bibfield{author}{\bibinfo{person}{Karan Ganju}, \bibinfo{person}{Qi Wang},
  \bibinfo{person}{Wei Yang}, \bibinfo{person}{Carl~A. Gunter}, {and}
  \bibinfo{person}{Nikita Borisov}.} \bibinfo{year}{2018}\natexlab{}.
\newblock \showarticletitle{Property Inference Attacks on Fully Connected
  Neural Networks using Permutation Invariant Representations}. In
  \bibinfo{booktitle}{\emph{Proceedings of the 2018 {ACM} {SIGSAC} Conference
  on Computer and Communications Security, {CCS} 2018, Toronto, ON, Canada,
  October 15-19, 2018}}. \bibinfo{publisher}{{ACM}}, \bibinfo{pages}{619--633}.
\newblock


\bibitem[Goodfellow et~al\mbox{.}(2015)]%
        {goodfellow2014explaining}
\bibfield{author}{\bibinfo{person}{Ian~J. Goodfellow},
  \bibinfo{person}{Jonathon Shlens}, {and} \bibinfo{person}{Christian
  Szegedy}.} \bibinfo{year}{2015}\natexlab{}.
\newblock \showarticletitle{Explaining and Harnessing Adversarial Examples}. In
  \bibinfo{booktitle}{\emph{3rd International Conference on Learning
  Representations, {ICLR} 2015, San Diego, CA, USA, May 7-9, 2015, Conference
  Track Proceedings}}.
\newblock


\bibitem[Goodman and Flaxman(2017)]%
        {goodman2017european}
\bibfield{author}{\bibinfo{person}{Bryce Goodman} {and} \bibinfo{person}{Seth
  Flaxman}.} \bibinfo{year}{2017}\natexlab{}.
\newblock \showarticletitle{European Union regulations on algorithmic
  decision-making and a “right to explanation”}.
\newblock \bibinfo{journal}{\emph{AI magazine}} \bibinfo{volume}{38},
  \bibinfo{number}{3} (\bibinfo{year}{2017}), \bibinfo{pages}{50--57}.
\newblock


\bibitem[Gregorutti et~al\mbox{.}(2017)]%
        {gregorutti2017correlation}
\bibfield{author}{\bibinfo{person}{Baptiste Gregorutti},
  \bibinfo{person}{Bertrand Michel}, {and} \bibinfo{person}{Philippe
  Saint-Pierre}.} \bibinfo{year}{2017}\natexlab{}.
\newblock \showarticletitle{Correlation and variable importance in random
  forests}.
\newblock \bibinfo{journal}{\emph{Statistics and Computing}}
  \bibinfo{volume}{27}, \bibinfo{number}{3} (\bibinfo{year}{2017}),
  \bibinfo{pages}{659--678}.
\newblock


\bibitem[Hornik et~al\mbox{.}(1989)]%
        {hornik1989multilayer}
\bibfield{author}{\bibinfo{person}{Kurt Hornik}, \bibinfo{person}{Maxwell
  Stinchcombe}, {and} \bibinfo{person}{Halbert White}.}
  \bibinfo{year}{1989}\natexlab{}.
\newblock \showarticletitle{Multilayer feedforward networks are universal
  approximators}.
\newblock \bibinfo{journal}{\emph{Neural networks}} \bibinfo{volume}{2},
  \bibinfo{number}{5} (\bibinfo{year}{1989}), \bibinfo{pages}{359--366}.
\newblock


\bibitem[Kairouz et~al\mbox{.}(2021)]%
        {kairouz2021advances}
\bibfield{author}{\bibinfo{person}{Peter Kairouz}, \bibinfo{person}{H~Brendan
  McMahan}, \bibinfo{person}{Brendan Avent}, \bibinfo{person}{Aur{\'e}lien
  Bellet}, \bibinfo{person}{Mehdi Bennis}, \bibinfo{person}{Arjun~Nitin
  Bhagoji}, \bibinfo{person}{Kallista Bonawitz}, \bibinfo{person}{Zachary
  Charles}, \bibinfo{person}{Graham Cormode}, \bibinfo{person}{Rachel
  Cummings}, {et~al\mbox{.}}} \bibinfo{year}{2021}\natexlab{}.
\newblock \showarticletitle{Advances and open problems in federated learning}.
\newblock \bibinfo{journal}{\emph{Foundations and Trends{\textregistered} in
  Machine Learning}} \bibinfo{volume}{14}, \bibinfo{number}{1--2}
  (\bibinfo{year}{2021}), \bibinfo{pages}{1--210}.
\newblock


\bibitem[Kim and Routledge(2018)]%
        {kim2018informational}
\bibfield{author}{\bibinfo{person}{Tae~Wan Kim} {and} \bibinfo{person}{Bryan~R.
  Routledge}.} \bibinfo{year}{2018}\natexlab{}.
\newblock \showarticletitle{Informational Privacy, {A} Right to Explanation,
  and Interpretable {AI}}. In \bibinfo{booktitle}{\emph{2018 {IEEE} Symposium
  on Privacy-Aware Computing, {PAC} 2018, Washington, DC, USA, September 26-28,
  2018}}. \bibinfo{publisher}{{IEEE}}, \bibinfo{pages}{64--74}.
\newblock


\bibitem[Lundberg and Lee(2017)]%
        {lundberg2017SHAP}
\bibfield{author}{\bibinfo{person}{Scott~M. Lundberg} {and}
  \bibinfo{person}{Su{-}In Lee}.} \bibinfo{year}{2017}\natexlab{}.
\newblock \showarticletitle{A Unified Approach to Interpreting Model
  Predictions}. In \bibinfo{booktitle}{\emph{Advances in Neural Information
  Processing Systems 30: Annual Conference on Neural Information Processing
  Systems 2017, December 4-9, 2017, Long Beach, CA, {USA}}}.
  \bibinfo{pages}{4765--4774}.
\newblock


\bibitem[Luo et~al\mbox{.}(2021)]%
        {luo2021feature}
\bibfield{author}{\bibinfo{person}{Xinjian Luo}, \bibinfo{person}{Yuncheng Wu},
  \bibinfo{person}{Xiaokui Xiao}, {and} \bibinfo{person}{Beng~Chin Ooi}.}
  \bibinfo{year}{2021}\natexlab{}.
\newblock \showarticletitle{Feature Inference Attack on Model Predictions in
  Vertical Federated Learning}. In \bibinfo{booktitle}{\emph{37th {IEEE}
  International Conference on Data Engineering, {ICDE} 2021, Chania, Greece,
  April 19-22, 2021}}. \bibinfo{publisher}{{IEEE}}, \bibinfo{pages}{181--192}.
\newblock


\bibitem[Luo et~al\mbox{.}(2022)]%
        {luo2021fusion}
\bibfield{author}{\bibinfo{person}{Xinjian Luo}, \bibinfo{person}{Xiaokui
  Xiao}, \bibinfo{person}{Yuncheng Wu}, \bibinfo{person}{Juncheng Liu}, {and}
  \bibinfo{person}{Beng~Chin Ooi}.} \bibinfo{year}{2022}\natexlab{}.
\newblock \showarticletitle{A Fusion-Denoising Attack on InstaHide with Data
  Augmentation}. In \bibinfo{booktitle}{\emph{Thirty-Sixth {AAAI} Conference on
  Artificial Intelligence, {AAAI} 2022, Virtual Event, February 22 - March 1,
  2022}}. \bibinfo{publisher}{{AAAI} Press}, \bibinfo{pages}{1899--1907}.
\newblock


\bibitem[Maleki et~al\mbox{.}(2013)]%
        {maleki2013bounding}
\bibfield{author}{\bibinfo{person}{Sasan Maleki}, \bibinfo{person}{Long
  Tran{-}Thanh}, \bibinfo{person}{Greg Hines}, \bibinfo{person}{Talal Rahwan},
  {and} \bibinfo{person}{Alex Rogers}.} \bibinfo{year}{2013}\natexlab{}.
\newblock \showarticletitle{Bounding the Estimation Error of Sampling-based
  Shapley Value Approximation}.
\newblock \bibinfo{journal}{\emph{CoRR}}  \bibinfo{volume}{abs/1306.4265}
  (\bibinfo{year}{2013}).
\newblock
\showeprint[arXiv]{1306.4265}
\urldef\tempurl%
\url{http://arxiv.org/abs/1306.4265}
\showURL{%
\tempurl}


\bibitem[McCaffrey(1992)]%
        {hastie2017generalized}
\bibfield{author}{\bibinfo{person}{Daniel~F. McCaffrey}.}
  \bibinfo{year}{1992}\natexlab{}.
\newblock \showarticletitle{Generalized Additive Models {(T.} J. Hastie and R.
  J. Tibshirani)}.
\newblock \bibinfo{journal}{\emph{{SIAM} Rev.}} \bibinfo{volume}{34},
  \bibinfo{number}{4} (\bibinfo{year}{1992}), \bibinfo{pages}{675--678}.
\newblock


\bibitem[Melis et~al\mbox{.}(2019)]%
        {melis2019exploiting}
\bibfield{author}{\bibinfo{person}{Luca Melis}, \bibinfo{person}{Congzheng
  Song}, \bibinfo{person}{Emiliano~De Cristofaro}, {and}
  \bibinfo{person}{Vitaly Shmatikov}.} \bibinfo{year}{2019}\natexlab{}.
\newblock \showarticletitle{Exploiting Unintended Feature Leakage in
  Collaborative Learning}. In \bibinfo{booktitle}{\emph{2019 {IEEE} Symposium
  on Security and Privacy, {SP} 2019, San Francisco, CA, USA, May 19-23,
  2019}}. \bibinfo{publisher}{{IEEE}}, \bibinfo{pages}{691--706}.
\newblock


\bibitem[Milli et~al\mbox{.}(2019)]%
        {milli2019model}
\bibfield{author}{\bibinfo{person}{Smitha Milli}, \bibinfo{person}{Ludwig
  Schmidt}, \bibinfo{person}{Anca~D. Dragan}, {and} \bibinfo{person}{Moritz
  Hardt}.} \bibinfo{year}{2019}\natexlab{}.
\newblock \showarticletitle{Model Reconstruction from Model Explanations}. In
  \bibinfo{booktitle}{\emph{Proceedings of the Conference on Fairness,
  Accountability, and Transparency, FAT* 2019, Atlanta, GA, USA, January 29-31,
  2019}}. \bibinfo{publisher}{{ACM}}, \bibinfo{pages}{1--9}.
\newblock


\bibitem[Mohassel and Zhang(2017)]%
        {mohassel2017secureml}
\bibfield{author}{\bibinfo{person}{Payman Mohassel} {and}
  \bibinfo{person}{Yupeng Zhang}.} \bibinfo{year}{2017}\natexlab{}.
\newblock \showarticletitle{SecureML: {A} System for Scalable
  Privacy-Preserving Machine Learning}. In \bibinfo{booktitle}{\emph{2017
  {IEEE} Symposium on Security and Privacy, {SP} 2017, San Jose, CA, USA, May
  22-26, 2017}}. \bibinfo{publisher}{{IEEE} Computer Society},
  \bibinfo{pages}{19--38}.
\newblock


\bibitem[Moro et~al\mbox{.}(2014)]%
        {bank}
\bibfield{author}{\bibinfo{person}{S{\'e}rgio Moro}, \bibinfo{person}{Paulo
  Cortez}, {and} \bibinfo{person}{Paulo Rita}.}
  \bibinfo{year}{2014}\natexlab{}.
\newblock \showarticletitle{A data-driven approach to predict the success of
  bank telemarketing}.
\newblock \bibinfo{journal}{\emph{Decision Support Systems}}
  \bibinfo{volume}{62} (\bibinfo{year}{2014}), \bibinfo{pages}{22--31}.
\newblock


\bibitem[Nasr et~al\mbox{.}(2019)]%
        {nasr2019comprehensive}
\bibfield{author}{\bibinfo{person}{Milad Nasr}, \bibinfo{person}{Reza Shokri},
  {and} \bibinfo{person}{Amir Houmansadr}.} \bibinfo{year}{2019}\natexlab{}.
\newblock \showarticletitle{Comprehensive Privacy Analysis of Deep Learning:
  Passive and Active White-box Inference Attacks against Centralized and
  Federated Learning}. In \bibinfo{booktitle}{\emph{2019 {IEEE} Symposium on
  Security and Privacy, {SP} 2019, San Francisco, CA, USA, May 19-23, 2019}}.
  \bibinfo{publisher}{{IEEE}}, \bibinfo{pages}{739--753}.
\newblock


\bibitem[Pilario et~al\mbox{.}(2019)]%
        {pilario2019review}
\bibfield{author}{\bibinfo{person}{Karl~Ezra Pilario}, \bibinfo{person}{Mahmood
  Shafiee}, \bibinfo{person}{Yi Cao}, \bibinfo{person}{Liyun Lao}, {and}
  \bibinfo{person}{Shuang-Hua Yang}.} \bibinfo{year}{2019}\natexlab{}.
\newblock \showarticletitle{A review of kernel methods for feature extraction
  in nonlinear process monitoring}.
\newblock \bibinfo{journal}{\emph{Processes}} \bibinfo{volume}{8},
  \bibinfo{number}{1} (\bibinfo{year}{2019}), \bibinfo{pages}{24}.
\newblock


\bibitem[R{\"a}tsch et~al\mbox{.}(2006)]%
        {ratsch2006learning}
\bibfield{author}{\bibinfo{person}{Gunnar R{\"a}tsch},
  \bibinfo{person}{S{\"o}ren Sonnenburg}, {and} \bibinfo{person}{Christin
  Sch{\"a}fer}.} \bibinfo{year}{2006}\natexlab{}.
\newblock \showarticletitle{Learning interpretable SVMs for biological sequence
  classification}. In \bibinfo{booktitle}{\emph{BMC bioinformatics}},
  Vol.~\bibinfo{volume}{7}. Springer, \bibinfo{pages}{1--14}.
\newblock


\bibitem[Reinders and Rosenhahn(2019)]%
        {reinders2019neural}
\bibfield{author}{\bibinfo{person}{Christoph Reinders} {and}
  \bibinfo{person}{Bodo Rosenhahn}.} \bibinfo{year}{2019}\natexlab{}.
\newblock \showarticletitle{Neural Random Forest Imitation}.
\newblock \bibinfo{journal}{\emph{CoRR}}  \bibinfo{volume}{abs/1911.10829}
  (\bibinfo{year}{2019}).
\newblock
\showeprint[arXiv]{1911.10829}
\urldef\tempurl%
\url{http://arxiv.org/abs/1911.10829}
\showURL{%
\tempurl}


\bibitem[Ribeiro et~al\mbox{.}(2016)]%
        {ribeiro2016LIME}
\bibfield{author}{\bibinfo{person}{Marco~Tulio Ribeiro},
  \bibinfo{person}{Sameer Singh}, {and} \bibinfo{person}{Carlos Guestrin}.}
  \bibinfo{year}{2016}\natexlab{}.
\newblock \showarticletitle{"Why Should {I} Trust You?" Explaining the
  predictions of any classifier}. In \bibinfo{booktitle}{\emph{Proceedings of
  the 22nd {ACM} {SIGKDD} International Conference on Knowledge Discovery and
  Data Mining, San Francisco, CA, USA, August 13-17, 2016}}.
  \bibinfo{publisher}{{ACM}}, \bibinfo{pages}{1135--1144}.
\newblock


\bibitem[Rudin(2019)]%
        {rudin2019stop}
\bibfield{author}{\bibinfo{person}{Cynthia Rudin}.}
  \bibinfo{year}{2019}\natexlab{}.
\newblock \showarticletitle{Stop explaining black box machine learning models
  for high stakes decisions and use interpretable models instead}.
\newblock \bibinfo{journal}{\emph{Nature Machine Intelligence}}
  \bibinfo{volume}{1}, \bibinfo{number}{5} (\bibinfo{year}{2019}),
  \bibinfo{pages}{206--215}.
\newblock


\bibitem[Sagi and Rokach(2018)]%
        {sagi2018ensemble}
\bibfield{author}{\bibinfo{person}{Omer Sagi} {and} \bibinfo{person}{Lior
  Rokach}.} \bibinfo{year}{2018}\natexlab{}.
\newblock \showarticletitle{Ensemble learning: A survey}.
\newblock \bibinfo{journal}{\emph{Wiley Interdisciplinary Reviews: Data Mining
  and Knowledge Discovery}} \bibinfo{volume}{8}, \bibinfo{number}{4}
  (\bibinfo{year}{2018}), \bibinfo{pages}{e1249}.
\newblock


\bibitem[Salem et~al\mbox{.}(2019)]%
        {salem2018ml}
\bibfield{author}{\bibinfo{person}{Ahmed Salem}, \bibinfo{person}{Yang Zhang},
  \bibinfo{person}{Mathias Humbert}, \bibinfo{person}{Pascal Berrang},
  \bibinfo{person}{Mario Fritz}, {and} \bibinfo{person}{Michael Backes}.}
  \bibinfo{year}{2019}\natexlab{}.
\newblock \showarticletitle{ML-Leaks: Model and Data Independent Membership
  Inference Attacks and Defenses on Machine Learning Models}. In
  \bibinfo{booktitle}{\emph{26th Annual Network and Distributed System Security
  Symposium, {NDSS} 2019, San Diego, California, USA, February 24-27, 2019}}.
  \bibinfo{publisher}{The Internet Society}.
\newblock


\bibitem[Selbst and Powles(2018)]%
        {selbst2018meaningful}
\bibfield{author}{\bibinfo{person}{Andrew Selbst} {and} \bibinfo{person}{Julia
  Powles}.} \bibinfo{year}{2018}\natexlab{}.
\newblock \showarticletitle{"Meaningful Information" and the Right to
  Explanation}. In \bibinfo{booktitle}{\emph{Conference on Fairness,
  Accountability and Transparency, {FAT} 2018, 23-24 February 2018, New York,
  NY, {USA}}}, Vol.~\bibinfo{volume}{81}. \bibinfo{publisher}{{PMLR}},
  \bibinfo{pages}{48}.
\newblock


\bibitem[Selvaraju et~al\mbox{.}(2017)]%
        {selvaraju2017grad}
\bibfield{author}{\bibinfo{person}{Ramprasaath~R. Selvaraju},
  \bibinfo{person}{Michael Cogswell}, \bibinfo{person}{Abhishek Das},
  \bibinfo{person}{Ramakrishna Vedantam}, \bibinfo{person}{Devi Parikh}, {and}
  \bibinfo{person}{Dhruv Batra}.} \bibinfo{year}{2017}\natexlab{}.
\newblock \showarticletitle{Grad-CAM: Visual Explanations from Deep Networks
  via Gradient-Based Localization}. In \bibinfo{booktitle}{\emph{{IEEE}
  International Conference on Computer Vision, {ICCV} 2017, Venice, Italy,
  October 22-29, 2017}}. \bibinfo{publisher}{{IEEE} Computer Society},
  \bibinfo{pages}{618--626}.
\newblock


\bibitem[Shapley(1953)]%
        {shapley1953}
\bibfield{author}{\bibinfo{person}{Lloyd~S Shapley}.}
  \bibinfo{year}{1953}\natexlab{}.
\newblock \showarticletitle{A value for n-person games}.
\newblock Vol.~\bibinfo{volume}{2}. \bibinfo{publisher}{Princeton University
  Press}, \bibinfo{pages}{303--317}.
\newblock


\bibitem[Shokri et~al\mbox{.}(2021)]%
        {shokri2021privacy}
\bibfield{author}{\bibinfo{person}{Reza Shokri}, \bibinfo{person}{Martin
  Strobel}, {and} \bibinfo{person}{Yair Zick}.}
  \bibinfo{year}{2021}\natexlab{}.
\newblock \showarticletitle{On the Privacy Risks of Model Explanations}. In
  \bibinfo{booktitle}{\emph{{AIES} '21: {AAAI/ACM} Conference on AI, Ethics,
  and Society, Virtual Event, USA, May 19-21, 2021}}.
  \bibinfo{publisher}{{ACM}}, \bibinfo{pages}{231--241}.
\newblock


\bibitem[Shokri et~al\mbox{.}(2017)]%
        {shokri2017membership}
\bibfield{author}{\bibinfo{person}{Reza Shokri}, \bibinfo{person}{Marco
  Stronati}, \bibinfo{person}{Congzheng Song}, {and} \bibinfo{person}{Vitaly
  Shmatikov}.} \bibinfo{year}{2017}\natexlab{}.
\newblock \showarticletitle{Membership Inference Attacks Against Machine
  Learning Models}. In \bibinfo{booktitle}{\emph{2017 {IEEE} Symposium on
  Security and Privacy, {SP} 2017, San Jose, CA, USA, May 22-26, 2017}}.
  \bibinfo{publisher}{{IEEE} Computer Society}, \bibinfo{pages}{3--18}.
\newblock


\bibitem[Shrikumar et~al\mbox{.}(2017)]%
        {shrikumar2017DeepLIFT}
\bibfield{author}{\bibinfo{person}{Avanti Shrikumar}, \bibinfo{person}{Peyton
  Greenside}, {and} \bibinfo{person}{Anshul Kundaje}.}
  \bibinfo{year}{2017}\natexlab{}.
\newblock \showarticletitle{Learning Important Features Through Propagating
  Activation Differences}. In \bibinfo{booktitle}{\emph{Proceedings of the 34th
  International Conference on Machine Learning, {ICML} 2017, Sydney, NSW,
  Australia, 6-11 August 2017}}, Vol.~\bibinfo{volume}{70}.
  \bibinfo{publisher}{{PMLR}}, \bibinfo{pages}{3145--3153}.
\newblock


\bibitem[Srivastava et~al\mbox{.}(2014)]%
        {srivastava2014dropout}
\bibfield{author}{\bibinfo{person}{Nitish Srivastava},
  \bibinfo{person}{Geoffrey Hinton}, \bibinfo{person}{Alex Krizhevsky},
  \bibinfo{person}{Ilya Sutskever}, {and} \bibinfo{person}{Ruslan
  Salakhutdinov}.} \bibinfo{year}{2014}\natexlab{}.
\newblock \showarticletitle{Dropout: a simple way to prevent neural networks
  from overfitting}.
\newblock \bibinfo{journal}{\emph{The journal of machine learning research}}
  \bibinfo{volume}{15}, \bibinfo{number}{1} (\bibinfo{year}{2014}),
  \bibinfo{pages}{1929--1958}.
\newblock


\bibitem[Strack et~al\mbox{.}(2014)]%
        {diabetes}
\bibfield{author}{\bibinfo{person}{Beata Strack}, \bibinfo{person}{Jonathan~P
  DeShazo}, \bibinfo{person}{Chris Gennings}, \bibinfo{person}{Juan~L Olmo},
  \bibinfo{person}{Sebastian Ventura}, \bibinfo{person}{Krzysztof~J Cios},
  {and} \bibinfo{person}{John~N Clore}.} \bibinfo{year}{2014}\natexlab{}.
\newblock \showarticletitle{Impact of HbA1c measurement on hospital readmission
  rates: analysis of 70,000 clinical database patient records}.
\newblock \bibinfo{journal}{\emph{BioMed research international}}
  \bibinfo{volume}{2014} (\bibinfo{year}{2014}).
\newblock


\bibitem[{\v{S}}trumbelj and Kononenko(2014)]%
        {vstrumbelj2014explaining}
\bibfield{author}{\bibinfo{person}{Erik {\v{S}}trumbelj} {and}
  \bibinfo{person}{Igor Kononenko}.} \bibinfo{year}{2014}\natexlab{}.
\newblock \showarticletitle{Explaining prediction models and individual
  predictions with feature contributions}.
\newblock \bibinfo{journal}{\emph{Knowledge and information systems}}
  \bibinfo{volume}{41}, \bibinfo{number}{3} (\bibinfo{year}{2014}),
  \bibinfo{pages}{647--665}.
\newblock


\bibitem[Tram{\`{e}}r et~al\mbox{.}(2016)]%
        {tramer2016stealing}
\bibfield{author}{\bibinfo{person}{Florian Tram{\`{e}}r}, \bibinfo{person}{Fan
  Zhang}, \bibinfo{person}{Ari Juels}, \bibinfo{person}{Michael~K. Reiter},
  {and} \bibinfo{person}{Thomas Ristenpart}.} \bibinfo{year}{2016}\natexlab{}.
\newblock \showarticletitle{Stealing Machine Learning Models via Prediction
  APIs}. In \bibinfo{booktitle}{\emph{25th {USENIX} Security Symposium,
  {USENIX} Security 16, Austin, TX, USA, August 10-12, 2016}}.
  \bibinfo{publisher}{{USENIX} Association}, \bibinfo{pages}{601--618}.
\newblock


\bibitem[Van Der~Putten and van Someren(2000)]%
        {insurance}
\bibfield{author}{\bibinfo{person}{Peter Van Der~Putten} {and}
  \bibinfo{person}{Maarten van Someren}.} \bibinfo{year}{2000}\natexlab{}.
\newblock \bibinfo{booktitle}{\emph{CoIL challenge 2000: The insurance company
  case}}.
\newblock \bibinfo{type}{{T}echnical {R}eport}. \bibinfo{institution}{Technical
  Report 2000--09, Leiden Institute of Advanced Computer Science}.
\newblock


\bibitem[Wang et~al\mbox{.}(2021)]%
        {wang2021privacy}
\bibfield{author}{\bibinfo{person}{Binghui Wang}, \bibinfo{person}{Jiayi Guo},
  \bibinfo{person}{Ang Li}, \bibinfo{person}{Yiran Chen}, {and}
  \bibinfo{person}{Hai Li}.} \bibinfo{year}{2021}\natexlab{}.
\newblock \showarticletitle{Privacy-Preserving Representation Learning on
  Graphs: {A} Mutual Information Perspective}. In
  \bibinfo{booktitle}{\emph{{KDD} '21: The 27th {ACM} {SIGKDD} Conference on
  Knowledge Discovery and Data Mining, Virtual Event, Singapore, August 14-18,
  2021}}. \bibinfo{publisher}{{ACM}}, \bibinfo{pages}{1667--1676}.
\newblock


\bibitem[{Wikipedia contributors}(2022a)]%
        {Chebyshev}
\bibfield{author}{\bibinfo{person}{{Wikipedia contributors}}.}
  \bibinfo{year}{2022}\natexlab{a}.
\newblock \bibinfo{title}{Chebyshev's inequality --- {Wikipedia}}.
\newblock
  \bibinfo{howpublished}{\url{https://en.wikipedia.org/wiki/Chebyshev\%27s_inequality}}.
\newblock
\newblock
\shownote{Online; accessed 3-April-2022}.


\bibitem[{Wikipedia contributors}(2022b)]%
        {Hoeffding}
\bibfield{author}{\bibinfo{person}{{Wikipedia contributors}}.}
  \bibinfo{year}{2022}\natexlab{b}.
\newblock \bibinfo{title}{Hoeffding's inequality --- {Wikipedia}}.
\newblock
  \bibinfo{howpublished}{\url{https://en.wikipedia.org/wiki/Hoeffding\%27s_inequality}}.
\newblock
\newblock
\shownote{[Online; accessed 3-April-2022]}.


\bibitem[{Wikipedia contributors}(2022c)]%
        {Pearson}
\bibfield{author}{\bibinfo{person}{{Wikipedia contributors}}.}
  \bibinfo{year}{2022}\natexlab{c}.
\newblock \bibinfo{title}{Pearson correlation coefficient}.
\newblock
  \bibinfo{howpublished}{\url{https://en.wikipedia.org/wiki/Pearson_correlation_coefficient}}.
\newblock
\newblock
\shownote{Online; accessed 1-April-2022}.


\bibitem[Wu et~al\mbox{.}(2016)]%
        {wu2016methodology}
\bibfield{author}{\bibinfo{person}{Xi Wu}, \bibinfo{person}{Matthew
  Fredrikson}, \bibinfo{person}{Somesh Jha}, {and} \bibinfo{person}{Jeffrey~F.
  Naughton}.} \bibinfo{year}{2016}\natexlab{}.
\newblock \showarticletitle{A Methodology for Formalizing Model-Inversion
  Attacks}. In \bibinfo{booktitle}{\emph{{IEEE} 29th Computer Security
  Foundations Symposium, {CSF} 2016, Lisbon, Portugal, June 27 - July 1,
  2016}}. \bibinfo{publisher}{{IEEE} Computer Society},
  \bibinfo{pages}{355--370}.
\newblock


\bibitem[Wu et~al\mbox{.}(2020)]%
        {wu2020privacy}
\bibfield{author}{\bibinfo{person}{Yuncheng Wu}, \bibinfo{person}{Shaofeng
  Cai}, \bibinfo{person}{Xiaokui Xiao}, \bibinfo{person}{Gang Chen}, {and}
  \bibinfo{person}{Beng~Chin Ooi}.} \bibinfo{year}{2020}\natexlab{}.
\newblock \showarticletitle{Privacy Preserving Vertical Federated Learning for
  Tree-based Models}.
\newblock \bibinfo{journal}{\emph{Proc. {VLDB} Endow.}} \bibinfo{volume}{13},
  \bibinfo{number}{11} (\bibinfo{year}{2020}), \bibinfo{pages}{2090--2103}.
\newblock


\bibitem[Xiao et~al\mbox{.}(2011)]%
        {xiao2010differential}
\bibfield{author}{\bibinfo{person}{Xiaokui Xiao}, \bibinfo{person}{Guozhang
  Wang}, {and} \bibinfo{person}{Johannes Gehrke}.}
  \bibinfo{year}{2011}\natexlab{}.
\newblock \showarticletitle{Differential Privacy via Wavelet Transforms}.
\newblock \bibinfo{journal}{\emph{{IEEE} Trans. Knowl. Data Eng.}}
  \bibinfo{volume}{23}, \bibinfo{number}{8} (\bibinfo{year}{2011}),
  \bibinfo{pages}{1200--1214}.
\newblock


\bibitem[Yang et~al\mbox{.}(2021)]%
        {yang2021gami}
\bibfield{author}{\bibinfo{person}{Zebin Yang}, \bibinfo{person}{Aijun Zhang},
  {and} \bibinfo{person}{Agus Sudjianto}.} \bibinfo{year}{2021}\natexlab{}.
\newblock \showarticletitle{GAMI-Net: An explainable neural network based on
  generalized additive models with structured interactions}.
\newblock \bibinfo{journal}{\emph{Pattern Recognition}}  \bibinfo{volume}{120}
  (\bibinfo{year}{2021}), \bibinfo{pages}{108192}.
\newblock


\bibitem[Yeh and Lien(2009)]%
        {credit}
\bibfield{author}{\bibinfo{person}{I-Cheng Yeh} {and} \bibinfo{person}{Che-hui
  Lien}.} \bibinfo{year}{2009}\natexlab{}.
\newblock \showarticletitle{The comparisons of data mining techniques for the
  predictive accuracy of probability of default of credit card clients}.
\newblock \bibinfo{journal}{\emph{Expert systems with applications}}
  \bibinfo{volume}{36}, \bibinfo{number}{2} (\bibinfo{year}{2009}),
  \bibinfo{pages}{2473--2480}.
\newblock


\bibitem[Yeung(2008)]%
        {yeung2008information}
\bibfield{author}{\bibinfo{person}{Raymond~W Yeung}.}
  \bibinfo{year}{2008}\natexlab{}.
\newblock \bibinfo{booktitle}{\emph{Information theory and network coding}}.
\newblock \bibinfo{publisher}{Springer Science \& Business Media}.
\newblock


\bibitem[Zhao et~al\mbox{.}(2021)]%
        {zhao2021exploiting}
\bibfield{author}{\bibinfo{person}{Xuejun Zhao}, \bibinfo{person}{Wencan
  Zhang}, \bibinfo{person}{Xiaokui Xiao}, {and} \bibinfo{person}{Brian~Y.
  Lim}.} \bibinfo{year}{2021}\natexlab{}.
\newblock \showarticletitle{Exploiting Explanations for Model Inversion
  Attacks}. In \bibinfo{booktitle}{\emph{2021 {IEEE/CVF} International
  Conference on Computer Vision, {ICCV} 2021, Montreal, QC, Canada, October
  10-17, 2021}}. \bibinfo{publisher}{{IEEE}}, \bibinfo{pages}{662--672}.
\newblock


\bibitem[Zhu et~al\mbox{.}(2019)]%
        {zhu2019deep}
\bibfield{author}{\bibinfo{person}{Ligeng Zhu}, \bibinfo{person}{Zhijian Liu},
  {and} \bibinfo{person}{Song Han}.} \bibinfo{year}{2019}\natexlab{}.
\newblock \showarticletitle{Deep Leakage from Gradients}. In
  \bibinfo{booktitle}{\emph{Advances in Neural Information Processing Systems
  32: Annual Conference on Neural Information Processing Systems 2019, NeurIPS
  2019, December 8-14, 2019, Vancouver, BC, Canada}}.
  \bibinfo{pages}{14747--14756}.
\newblock


\end{thebibliography}

%
%
%
%
%
%
%
%

\end{document}